\documentclass[letterpaper]{article} 
\usepackage{aaai2026}  
\usepackage{times}  
\usepackage{helvet}  
\usepackage{courier}  
\usepackage[hyphens]{url}  
\usepackage{graphicx} 
\usepackage{natbib}  
\usepackage{caption} 
\usepackage{algorithm}
\usepackage{algorithmic}

\usepackage{newfloat}
\usepackage{listings}
\DeclareCaptionStyle{ruled}{labelfont=normalfont,labelsep=colon,strut=off} 
\floatstyle{ruled}
\newfloat{listing}{tb}{lst}{}
\floatname{listing}{Listing}
\usepackage{amssymb}
\usepackage{amsmath}
\usepackage{multirow}
\usepackage{makecell}
\usepackage{colortbl} 
\usepackage{gensymb}
\usepackage{marvosym}

\title{TrackGS: Optimizing COLMAP-Free 3D Gaussian Splatting \\with Global Track Constraints}
\author {
    Dongbo Shi\textsuperscript{\rm 1}\equalcontrib,
    Shen Cao\textsuperscript{\rm 2}\equalcontrib,
    Lubin Fan\textsuperscript{\rm 2\Letter},
    Bojian Wu\textsuperscript{\rm 2},
    Jinhui Guo\textsuperscript{\rm 2},
    Ligang Liu\textsuperscript{\rm 1},
    Renjie Chen\textsuperscript{\rm 1\Letter},
}
\affiliations {
    \textsuperscript{\rm 1}University of Science and Technology of China\\
    \textsuperscript{\rm 2}Independent Researcher\\
}

\begin{document}

\maketitle

\begin{abstract}
We present TrackGS, a novel method to integrate global feature tracks with 3D Gaussian Splatting (3DGS) for COLMAP-free novel view synthesis. While 3DGS delivers impressive rendering quality, its reliance on accurate precomputed camera parameters remains a significant limitation. Existing COLMAP-free approaches depend on local constraints that fail in complex scenarios. Our key innovation lies in leveraging feature tracks to establish global geometric constraints, enabling simultaneous optimization of camera parameters and 3D Gaussians. Specifically, we: (1) introduce track-constrained Gaussians that serve as geometric anchors, (2) propose novel 2D and 3D track losses to enforce multi-view consistency, and (3) derive differentiable formulations for camera intrinsics optimization. Extensive experiments on challenging real-world and synthetic datasets demonstrate state-of-the-art performance, with much lower pose error than previous methods while maintaining superior rendering quality. Our approach eliminates the need for COLMAP preprocessing, making 3DGS more accessible for practical applications.
\end{abstract}

\section{Introduction}
\label{sec:intro}
Given a collection of images from a 3D scene along with the corresponding camera intrinsic and extrinsic parameters, 3D Gaussian Splatting (3DGS)~\cite{3DGS2023} can effectively represent the scene with a series of 3D Gaussians, and generate high-quality images from novel viewpoints. Due to its efficiency in training and superior performance in testing, 3DGS has become popular for a variety of applications including reconstruction, editing, and AR/VR etc. However, the effectiveness of 3DGS training relies on accurately pre-determined camera poses (i.e., camera extrinsics) and camera focal lengths (i.e., camera intrinsics). These parameters are typically derived using COLMAP~\cite{colmap2016} in advance. This preprocessing step is not only time-consuming but also impacts the training performance of 3DGS, particularly when dealing with complex camera movements and scenes.

Recent COLMAP-Free approaches~\cite{iNerf2021,SCNeRF2021,bian2022nopenerf,tracknerf2024eccv,fan2024instantsplat,CF-3DGS-2024,ji2024sfmfree3dgaussiansplatting} have tried to eliminate COLMAP dependencies by adding local constraints, such as photometric consistency or sequential frame alignment. 
While these methods work well for simple scenes with smooth camera trajectories, they struggle with more challenging cases—such as wide-baseline views, rapid camera movements, or unordered image collections—due to their reliance on incremental optimization and local geometric cues. This often leads to pose drift, scale ambiguity, and degraded rendering quality.

To address this problem, we propose \textbf{TrackGS}, a novel framework that integrates global track information into 3DGS to enforce multi-view geometric consistency robustly.
Our key insight is that feature tracks provide strong, long-range constraints across disparate views, enabling stable optimization of both camera parameters and 3D Gaussians even in complex scenarios. 
Specifically, we introduce: 1. \textbf{Track Gaussians}, where a subset of Gaussians are anchored to 3D track points, ensuring their spatial accuracy through reprojection and backprojection losses.
2. \textbf{A fully differentiable pipeline}, that jointly optimizes camera intrinsics, extrinsics, and 3DGS parameters. We derive gradients for intrinsic parameters (e.g., focal length), enabling end-to-end training without any precomputed camera priors.
Experimental results on both public and synthetic datasets demonstrate that our joint optimization framework with global track constraints outperforms previous methods.

In summary, our contributions are as follows:
\begin{itemize}
    \item We propose a method to integrate track information with 3DGS, using global geometric constraints to simultaneously optimize camera parameters and 3DGS. To achieve this, we introduce 2D and 3D track losses to constrain reprojection and backprojection errors.
    \item We propose a joint optimization framework for all camera parameters and 3D Gaussians. Without relying on any known camera parameters, we achieve full differentiability for the entire pipeline, seamlessly integrating camera parameters estimation, including both intrinsics and extrinsics, with 3DGS training.
    \item On both {challenging} public and synthetic datasets, our approach outperforms previous methods on both camera parameters estimation and novel view synthesis.
\end{itemize}

\begin{figure*}[!t]
    \centering
    \includegraphics[width=0.9\linewidth]{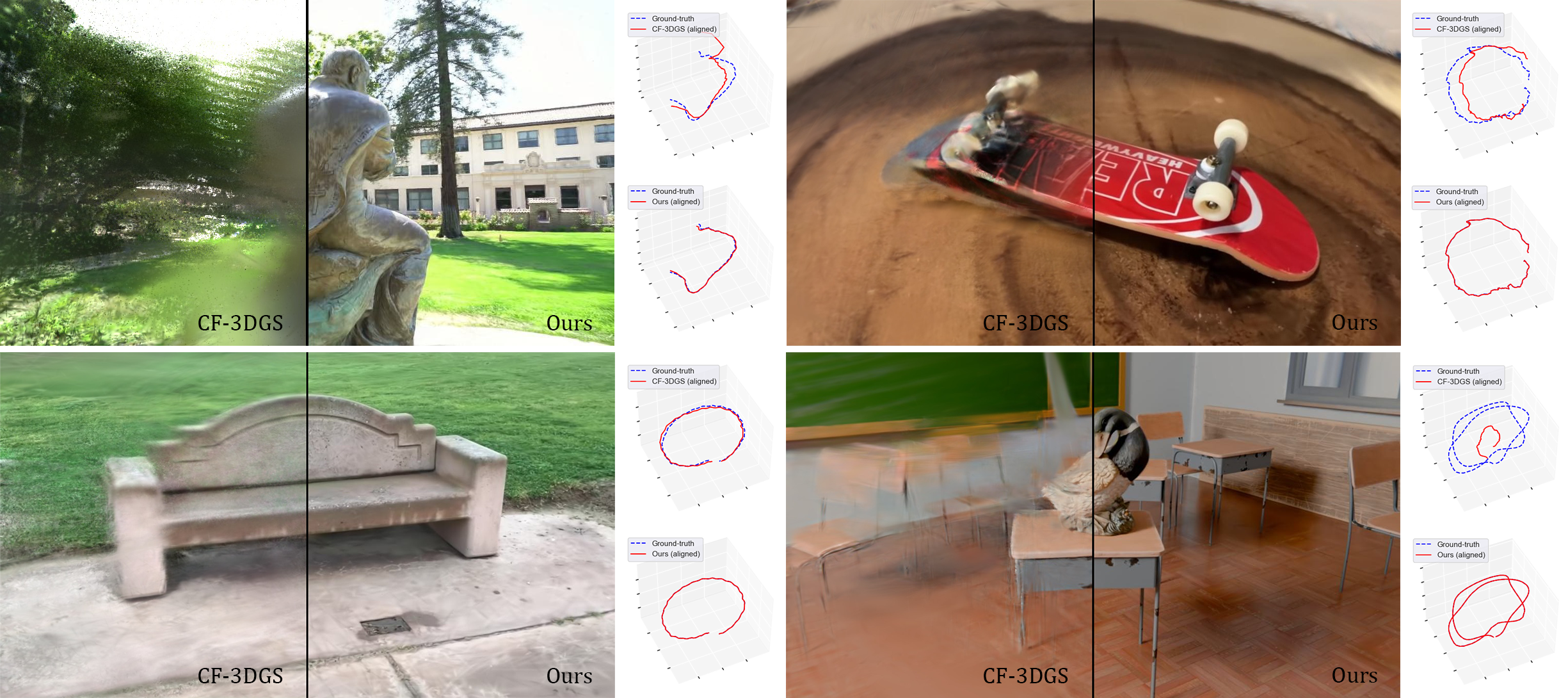}
    \caption{\textbf{Comparisons on novel view synthesis and camera poses.} We propose a novel 3DGS model without any known camera parameters by leveraging global track information. Compared with state-of-the-art methods, we provide higher rendering quality in novel view synthesis, and more accurate estimation of camera poses on benchmark datasets, including the {challenging} real-world indoor and outdoor or synthetic scenes with \textit{complicated} camera movements (the right column).}
    \label{fig:teaser}
\end{figure*}

\section{Related Work}
\label{sec:related-works}
\subsection{Novel View Synthesis}
Novel view synthesis is a foundational task in the computer vision and graphics, which aims to generate unseen views of a scene from a given set of images.
Numerous methods have been developed to address this problem by approaching it as 3D geometry-based rendering, such as using meshes~\cite{worldsheet,FVS,SVS}, MPI~\cite{MINE,single_view_mpi,stereo_magnification}, point clouds~\cite{point_differentiable,point_nfs}, etc.

\begin{figure*}[!t]
    \centering
    \includegraphics[width=0.9\linewidth]{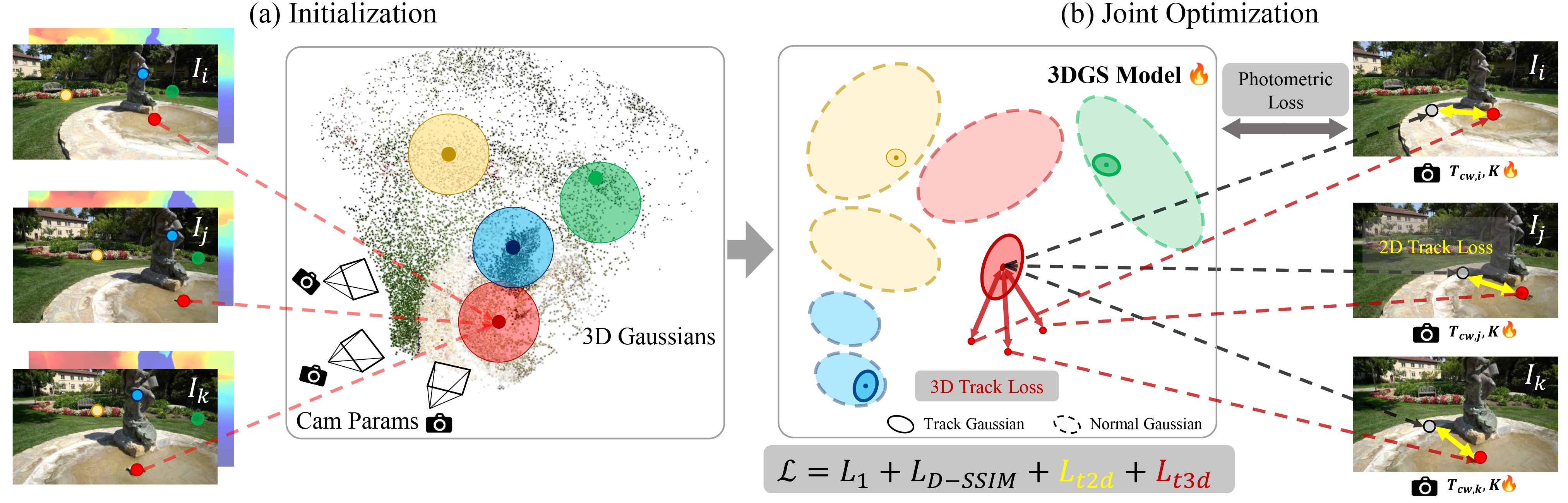}
    \caption{\textbf{Overview.} 
    Given a set of images, our method obtains both camera intrinsics and extrinsics, as well as a 3DGS model. During the initialization, we extract the global tracks, and initialize camera parameters and Gaussians from image correspondences and monodepth. We determine Gaussian kernels with 3D track points, and then jointly optimize the parameters $K$, $T_{cw}$, 3DGS through the proposed global track constraints ($L_{t2d}$, $L_{t3d}$) and original photometric losses ($L_1$, $L_{D-SSIM}$).
    }
    \label{fig:overview}
\end{figure*}

Recently, Neural Radiance Fields (NeRF)~\cite{2020NeRF} provide a novel solution to this problem by representing scenes as implicit radiance fields using neural networks, achieving photo-realistic rendering quality. Although having some works in improving efficiency~\cite{instant_nerf2022, lin2022enerf}, the time-consuming training and rendering still limit its practicality.
Alternatively, 3D Gaussian Splatting (3DGS)~\cite{3DGS2023} models the scene as explicit Gaussian kernels, with differentiable splatting for rendering. Its improved real-time rendering performance, lower storage and efficiency, quickly attract more attentions.

\subsection{Optimizing Camera Poses in NeRFs and 3DGS}
Although NeRF and 3DGS can provide impressive scene representation, these methods all need accurate camera parameters (both intrinsic and extrinsic) as additional inputs, which are mostly obtained by COLMAP~\cite{colmap2016}.
When the prior is inaccurate or unknown, accurately estimating camera parameters and scene representations becomes crucial.

In earlier studies, scene training and camera pose estimation relied solely on photometric constraints. iNeRF~\cite{iNerf2021} refines the camera poses using a pre-trained NeRF model. NeRFmm~\cite{wang2021nerfmm} introduces a joint optimization approach that simultaneously estimates camera poses and trains the NeRF model. BARF~\cite{barf2021} and GARF~\cite{2022GARF} propose a new positional encoding strategy to address the gradient inconsistency issues in positional embedding, achieving promising results. However, these methods only yield satisfactory optimization when the initial pose is very close to the ground truth, as photometric constraints alone can only enhance camera estimation quality within a limited range. Subsequently, 
SC-NeRF~\cite{SCNeRF2021} minimizes a projected ray distance loss based on correspondence between adjacent frames. NoPe-NeRF~\cite{bian2022nopenerf} utilizes monocular depth maps as geometric priors and defines undistorted depth loss and relative pose constraints.

Regarding 3D Gaussian Splatting, CF-3DGS~\cite{CF-3DGS-2024} utilizes mono-depth information to refine the optimization of local 3DGS for relative pose estimation and subsequently learns a global 3DGS in a sequential manner. InstantSplat~\cite{fan2024instantsplat} targets sparse view scenes, initially employing MASt3R~\cite{mast3r_eccv24} to create a dense, pixel-aligned point set for initializing 3D Gaussian models and implements a parallel grid partitioning strategy to accelerate joint optimization. Jiang et~al.~\cite{Jiang_2024sig} develops an incremental method for reconstructing camera poses and scenes, but struggles with complex scenes and unordered images. 
SFGS~\cite{ji2024sfmfree3dgaussiansplatting} interpolates frames for training and splits the scene into local clips, using a hierarchical strategy to build 3DGS model. It works well for simple scenes, but fails with dramatic motions due to unstable interpolation and low efficiency.

However, most existing methods generally depend on sequentially ordered image inputs and incrementally optimize camera parameters and 3DGS, which often leads to drift errors and hinders achieving globally consistent results. Our work seeks to overcome these limitations.

\section{3D Gaussian Splatting}\label{subsec:3dgs}
3DGS models a scene using a set of 3D anisotropic Gaussians. Each Gaussian is parameterized by a centroid $\mu\in\mathbb{R}^{3}$, a quaternion factor $q\in\mathbb{R}^{4}$, a scale factor $s\in\mathbb{R}^{3}$, spherical harmonics (SH) coefficients of color $c\in\mathbb{R}^{k}$, and opacity $\alpha\in\mathbb{R}$. 
Donating the rotation matrix of quaternion $q$ and scale matrix of $s$ by $R\in\mathbb{R}^{3\times 3}$ and $S=\mathrm{diag}(s)$, the covariance matrix $\Sigma$ and Gaussian function $G(x)$ are:
\begin{equation}\label{eq:cov}
    \Sigma=RSS^\top R^\top, G(x)=\exp{(-\frac{1}{2}(x-\mu)^\top\Sigma^{-1}(x-\mu))}.
\end{equation}
Denoting projection matrix $T_{cw}=[R_{cw}|t_{cw}]$, which transforms points from the \textit{world} to \textit{camera} coordinate space, an image rendered from the specified view can be obtained as follows. First, the covariance matrix in camera coordinates $\Sigma^{\mathrm{2D}}$ is obtained by approximating the projection of 3D Gaussian in pixel coordinates, and can be expressed as:
\begin{equation}\label{eq:cov-2d}
    \Sigma^{\mathrm{2D}}=JR_{cw}\Sigma R_{cw}^\top J^\top,
\end{equation}
where $J$ is the Jacobian of the affine approximation of the projective transformation. The final rendered color $\hat{C}$ can be denoted as the alpha-blending of $N$ ordered Gaussians:
\begin{equation}
    \label{eq:3dgs-color}    
    \hat{C}=\sum_{i}^{N}c_{i}\alpha_{i}\prod_{j}^{i-1}(1-\alpha_{j}),
\end{equation}
where $c_{i}$ and $\alpha_{i}$ are the color and opacity of the Gaussians. 
Similarly, the depth of the scene perceived of a pixel is,
\begin{equation}
    \label{eq:3dgs-depth}
    \hat{D}=\sum_{i}^{N}d_{i}\alpha_{i}\prod_{j}^{i-1}(1-\alpha_{j}),
\end{equation}
where $d_i$ denotes the z-axis coordinate for the transformed Gaussian centers in the camera space.

Usually, the parameters of 3D Gaussians are optimized by rendering and comparing the rendered images with the ground-truths. The loss function $\mathcal{L}$ is defined as follows:
\begin{equation}\label{eq:loss_3dgs}
    \mathcal{L} = (1-\lambda)L_{1} + \lambda L_{\mathrm{D-SSIM}}.
\end{equation}
Typically, 3D Gaussians are initialized with Structure from Motion (SfM) point clouds obtained from the input images. 

\section{Method}
\textbf{Overview.}
Given a set of images $\mathcal{I}=\{I_{i}\}_{i=1}^{M}$, with unknown extrinsic matrix $T_{cw,i}$ at each view and unknown intrinsic matrix denoted by $K$, our method aims to build a 3D Gaussian Splatting (3DGS) model while simultaneously estimating both the extrinsic and intrinsic matrices, as shown in Fig.~\ref{fig:overview}. To achieve this goal, our key approach is to leverage the global track constraint to explicitly capture and enforce multi-view geometric consistency, which serves as the foundation for accurately estimating both the 3DGS model and the camera parameters.
During initialization, we extract 2D feature correspondences and derive global tracks using off-the-shelf methods. 
We then initialize both the camera parameters and the original 3D Gaussians with the estimated 3D track points. 
Subsequently, we extract a subset of Gaussian kernels from the vanilla model, referred to as \textit{track Gaussians}, to integrate the track-based geometric prior into 3DGS.  
Building on this, we further propose a joint optimization framework with two additional loss terms: a 2D track loss and a 3D track loss. 
The 2D track loss enforces multi-view geometric consistency by minimizing the error between the reprojected track pixels and their references. 
The 3D track loss constrains the track Gaussians to remain aligned with the scene surface by penalizing the distance between the back-projected 3D points (from rendered depth) and the track Gaussians.
We derive and implement the differentiable components of the camera parameters, including both the extrinsic and intrinsic matrices. 
This allows us to apply the chain rule, enabling seamless joint optimization of the 3DGS model and the camera parameters.

\subsection{Initialization}\label{subsec:init}
Since the training set contains only images, we need to extract essential information during initialization. 
To initialize the 3D Gaussians, we estimate monocular depth maps $D$ for each image $I$ using depth estimators such as DPT~\cite{dpt2021iccv}, following the setups in CF-3DGS~\cite{CF-3DGS-2024} and SFGS~\cite{ji2024sfmfree3dgaussiansplatting}. In addition, to construct global track constraints for later optimization, we detect 2D feature points $\{p_{i}\}$ in each image $I$ and compute feature correspondences across all images using off-the-shelf methods~\cite{detone2018superpoint,sarlin20superglue}.

\textbf{Camera Parameters.}
We assume all cameras share a standard pinhole model with no distortion, and the principal point locates at the center of the image, then the intrinsic matrix $K$ of camera is:
\begin{equation}\label{eq:intrinsic}
    K=\begin{bmatrix}
        f_{x} & 0 & c_{x} \\
        0 & f_{y} & c_{y} \\
        0 & 0 & 1
    \end{bmatrix},
\end{equation}
where $(c_{x},c_{y})$ is the principal point and $(f_{x},f_{y})$ is focal length. 
Empirically, we initialize the focal length with a field of view (FoV) of $60^\circ$ as:
\begin{equation}
    f_{x}=f_{y}=\frac{\sqrt{c_{x}^{2}+c_{y}^{2}}}{\tan(\mathrm{FoV}/2)}.
\end{equation}
Given the estimated mono-depth maps $D_i, D_j$ and 2D correspondences $p_i, p_j$, the relative transformation $T_{ij}$ between frames $i$ and $j$ is computed by aligning point clouds $p_{i}^{*}, p_{j}^{*}$, where $p^{*}=D(p)\cdot K^{-1} \cdot p$. 
These alignments provide a coarse initialization of both camera intrinsics and extrinsics.

\textbf{Global Tracks.}
By using Union-Find algorithm over feature points, We extract a set of tracks $\mathcal{P}$, where each element $(P,\{p_{i}\}_{i=1}^{l})\in\mathcal{P}$ represents a 3D track point $P$ and its corresponding matching points $\{p_{i}\}_{i=1}^{l}$ associated with the training images. With the estimated camera parameters and mono-depth maps, the 3D point $P$ is initialized as the average position of $\{p_{i}^{*}\}_{i}$.
Notably, we use track points solely to initialize 3D Gaussians, as their positions will be refined by global optimization and constraints to accurately represent object surfaces.

\subsection{Joint Optimization}

\subsubsection{Track Gaussians and Global Track Constraints.}
During joint optimization, we leverage the global tracks obtained during initialization to enforce multi-view geometric consistency in both 2D and 3D space. 
Our method is built on two key components. 
First, to associate the tracks with the 3DGS model, we select a subset of 3D Gaussians associated with each track, referred to as \textit{track Gaussians}, whose centroids coincide with the 3D track points and serve as geometric anchors within the model.
Second, we introduce two additional loss terms to impose global track constraints over these track Gaussians. 
The reprojection loss (2D track loss) enforces that the reprojections of 3D track points on each image closely aligned with the corresponding 2D feature points. 
The backprojection loss (3D track loss) enforces that the matched 2D feature points, when back-projected using the 3DGS rendering depth, remain close to the same corresponding 3D track points across different views. 
Beyond enforcing multi-view consistency, the backprojection loss further ensures that the track Gaussians treated as geometric anchors, as the projected feature points with rendering depth are considered to lie on the scene surface.
These points serve as key elements for optimizing camera parameters and enhancing the global geometric consistency.

\textbf{2D Track Loss.}
We reproject the 3D track point $P$ into the corresponding images using the associated camera parameters and compute the reprojection loss, which will be summed to calculate the total 2D track loss:
\begin{equation}
    L_{t2d} = \sum_{P\in\mathcal{P}}\frac{1}{l}{\sum_{i=1}^{l}\|p_{i}-K\cdot T_{cw,i} \cdot P\|}.
\end{equation}

\textbf{3D Track Loss.}
We backproject the 2D feature points into 3D scene using the rendered depth and camera parameters associated with each point. The backprojection error is then computed with respect to the 3D track point $P$. Then errors are aggregated to calculate the overall 3D track loss:
\begin{equation}
    L_{t3d}=\sum_{P\in\mathcal{P}}\frac{1}{l}{\sum_{i=1}^{l}\|d(p_i) \cdot T_{cw,i}^{-1} \cdot K^{-1} \cdot p_{i} - P\|},
\end{equation}
where $d(p_i)$ denotes the depth perceived from $p_i$ according to Eq.~\ref{eq:3dgs-depth}. 
Note, the 2D track loss relates to the track Gaussians that are mainly used for the optimization of camera parameters, whereas the 3D track loss requires the 3DGS rendered depth values during computation. This indirectly ties the optimization of the 3D track loss to the optimization of the 3DGS model and enhances the capability of multi-view geometric consistency. They are fundamentally different.

\textbf{Overall Objectives.} Combined with Eq.~\ref{eq:loss_3dgs}, our joint optimization can be formulated as:

\begin{equation}
    \mathcal{L} = (1-\lambda)L_{1}+\lambda L_{\mathrm{D-SSIM}}+\lambda_{t2d}L_{t2d}+\lambda_{t3d}L_{t3d}.
    \label{eq:overall}
\end{equation}

\subsubsection{Optimizing Camera Parameters.}
To optimize the camera parameters of 3D Gaussians simultaneously, the gradient of the loss function $\mathcal{L}$ with respect to the camera parameters are needed. We derive these gradients accordingly, where the gradient of extrinsic parameters is:
\begin{equation}
    \frac{\partial \mathcal{L}}{\partial T_{cw}}=\frac{\partial \mathcal{L}}{\partial t}q^\top,
\end{equation}
where $q=[\mu,1]^T$ and $t=T_{cw}q=[t_{x},t_{y},t_{z},t_{w}]^T$. 
Further, let $(\mu^{'},\Sigma^{'})$ be the 2D projection of the centroid and covariance $(\mu, \Sigma)$, the gradient of $\mathcal{L}$ respect to focal length $F=(f_{x},f_{y})$ can be computed, where $T=JR_{cw}$, as:
\begin{equation}
    \left\{
    \begin{aligned}
        & \frac{\partial \mathcal{L}}{\partial f_{x}} = \frac{t_x}{t_z}\frac{\partial \mathcal{L}}{\partial \mu_{x}^{'}} + <\frac{\partial \mathcal{L}}{\partial T}{R_{cw}^\top},\frac{\partial J}{\partial f_{x}}>, \\
        & \frac{\partial \mathcal{L}}{\partial f_{y}} = \frac{t_y}{t_z}\frac{\partial \mathcal{L}}{\partial \mu_{y}^{'}} + <\frac{\partial \mathcal{L}}{\partial T}{R_{cw}^\top},\frac{\partial J}{\partial f_{y}}>. \\
    \end{aligned}
    \right.
    \label{eq:grad_intrinsic}
\end{equation}
Please refer to the supplementary materials for more details.

\begin{figure*}[!t]
  \centering
   \includegraphics[width=\linewidth]{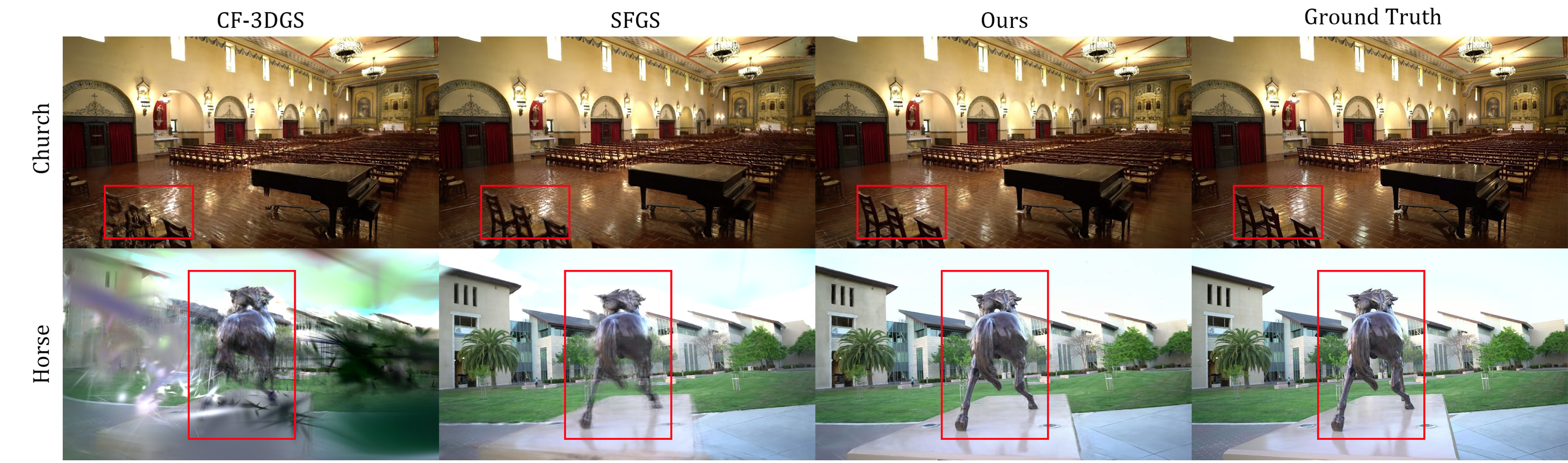}
   \caption{\textbf{Qualitative comparison for NVS on Tanks and Temples.}
   We achieve better rendering results on details.
   }
   \label{fig:render-T&T}
\end{figure*}

\begin{table*}[!t]
\setlength{\tabcolsep}{1mm}
    \centering
    {\fontsize{8}{9}\selectfont
    \begin{tabular}{c cccc cccc cccc ccc} \hline
         \multirow{2}{*}{T\&T} & \multicolumn{3}{c}{Church}  & & \multicolumn{3}{c}{Horse} & & \multicolumn{3}{c}{Ignatius} & & \multicolumn{3}{c}{Ballroom} \\ \cline{2-4}\cline{6-8}\cline{10-12}\cline{14-16}
                                 & PSNR$\uparrow$ & SSIM$\uparrow$ & LPIPS$\downarrow$ & & PSNR & SSIM & LPIPS & & PSNR & SSIM & LPIPS & & PSNR & SSIM & LPIPS   \\ \hline
    
         CF-3DGS    & 21.96 & 0.71 & 0.27   & & 17.36 & 0.64 & 0.33     & & 14.96 & 0.31 & 0.59     & & 22.20 & 0.72 & 0.25 \\
         SFGS       & 22.46 & 0.72 & 0.22   & & 20.86 & 0.75 & 0.19     & & 12.16 & 0.32 & 0.72     & & 17.42 & 0.49 & 0.27 \\
         Ours       & \cellcolor{gray!40}25.56& \cellcolor{gray!40}0.84& \cellcolor{gray!40}0.16    
                  & & \cellcolor{gray!40}27.60& \cellcolor{gray!40}0.90& \cellcolor{gray!40}0.12
                  & & \cellcolor{gray!40}22.12& \cellcolor{gray!40}0.71& \cellcolor{gray!40}0.22
                  & & \cellcolor{gray!40}25.94& \cellcolor{gray!40}0.86& \cellcolor{gray!40}0.12 \\
         \hline
         \multirow{2}{*}{CO3D V2} & \multicolumn{3}{c}{{46\_2587\_7531}}  & & \multicolumn{3}{c}{{110\_13051\_23361}} & & \multicolumn{3}{c}{{245\_26182\_52130}} & & \multicolumn{3}{c}{{415\_57112\_110099}} \\ \cline{2-4}\cline{6-8}\cline{10-12}\cline{14-16}
                    & PSNR & SSIM & LPIPS & & PSNR & SSIM & LPIPS & & PSNR & SSIM & LPIPS & & PSNR & SSIM & LPIPS   \\ \hline
         CF-3DGS    & 25.44 & 0.80 & 0.21   & & 29.69 & 0.89 & 0.29     & & 27.24 & 0.85 & 0.30     & & 22.14 & 0.64 & 0.34 \\
         SFGS       & 30.65 & 0.91 & 0.13   & & 29.95 & 0.87 & 0.19     & & 28.59 & 0.87 & 0.27     & & 27.23 & 0.78 & 0.30 \\
         Ours       & \cellcolor{gray!40}31.83 & \cellcolor{gray!40}0.92 & \cellcolor{gray!40}0.12    
                  & & \cellcolor{gray!40}33.44 & \cellcolor{gray!40}0.94 & \cellcolor{gray!40}0.11
                  & & \cellcolor{gray!40}33.82 & \cellcolor{gray!40}0.93 & \cellcolor{gray!40}0.20
                  & & \cellcolor{gray!40}30.37 & \cellcolor{gray!40}0.88 & \cellcolor{gray!40}0.22 \\
         \hline
         \multirow{2}{*}{Synthetic} & \multicolumn{3}{c}{classroom}  & & \multicolumn{3}{c}{{lego\_c2}} & & \multicolumn{3}{c}{livingroom} & & \multicolumn{3}{c}{bedroom} \\ \cline{2-4}\cline{6-8}\cline{10-12}\cline{14-16}
                    & PSNR & SSIM & LPIPS & & PSNR & SSIM & LPIPS & & PSNR & SSIM & LPIPS & & PSNR & SSIM & LPIPS   \\ \hline
         CF-3DGS    & 19.69 & 0.69 & 0.46   & & 15.93 & 0.31 & 0.55     & & 16.63 & 0.57 & 0.57     & & 16.98 & 0.65 & 0.45 \\
         SFGS       & 16.79 & 0.67 & 0.55   & & 13.64 & 0.29 & 0.68     & & 15.11 & 0.49 & 0.60     & & 12.39 & 0.54 & 0.54 \\
         Ours       & \cellcolor{gray!40}36.26 & \cellcolor{gray!40}0.94 & \cellcolor{gray!40}0.15    
                  & & \cellcolor{gray!40}29.36 & \cellcolor{gray!40}0.90 & \cellcolor{gray!40}0.12
                  & & \cellcolor{gray!40}33.52 & \cellcolor{gray!40}0.88 & \cellcolor{gray!40}0.24
                  & & \cellcolor{gray!40}31.17 & \cellcolor{gray!40}0.93 & \cellcolor{gray!40}0.13 \\
         \hline
    \end{tabular}
    }
    \caption{\textbf{NVS results on Tanks and Temples, CO3D-V2 and Synthetic.} Each baseline method is trained with its public code under the original settings and evaluated with the same evaluation protocol. 
    The best results are gray background.}
    \label{table:render-T&T-co3dv2-syn}
\end{table*}

\section{Experiments}\label{sec:results}

\subsection{Experimental Setup}
\textbf{Datasets.}
We conduct experiments on two real-world datasets, \emph{CO3D-V2}~\cite{co3d} and \emph{Tanks and Temples}~\cite{Tanks_Temples}, and a \emph{Synthetic Dataset} created by ourselves.
\textbf{CO3D-V2} includes thousands of videos of various objects. Following CF-3DGS~\cite{CF-3DGS-2024}, we select scenes with significant camera movements to demonstrate our robustness.
\textbf{Tanks and Temples} used in CF-3DGS is overly smooth and lacks camera motions which are common in real-world scenarios. We reduce the original 20 fps to 4 fps and perform sampling along a longer camera trajectory to obtain more challenging cases with large camera motions. 
\textbf{Synthetic Dataset} comprises 4 scenes with about 150 frames each created using Blender~\cite{blender2018}, showcasing complex roaming and object-centric camera motions. It's used to assess camera parameter estimation, providing ground truth for intrinsic and extrinsic parameters. For additional details on Tanks and Temples and Synthetic Dataset, see the supplementary material.

\textbf{Metrics.}
We use standard evaluation metrics, including PSNR, SSIM~\cite{SSIM}, and LPIPS~\cite{LPIPS} to evaluate the quality of novel view synthesis (NVS). For pose estimation, we rely on the Absolute Trajectory Error (ATE) and Relative Pose Error (RPE)~\cite{barf2021,bian2022nopenerf,CF-3DGS-2024}. $\mathrm{RPE}_r$ and $\mathrm{RPE}_t$ are utilized to measure the accuracy of rotation and translation. To ensure the metrics are comparable on the same scale, we align the camera poses using Umeyama's method~\cite{Umeyama} for both estimation and evaluation. For camera focal length, we convert it to the field of view (FoV) and calculate the angular error, following~\cite{zhu2023tame}.

\begin{table*}[!t]
\setlength{\tabcolsep}{1mm}
    \centering
    {\fontsize{8}{9}\selectfont
    \begin{tabular}{c cccc cccc cccc ccc} \hline
         \multirow{2}{*}{CO3D V2} & \multicolumn{3}{c}{{46\_2587\_7531}}  & & \multicolumn{3}{c}{{110\_13051\_23361}} & & \multicolumn{3}{c}{{245\_26182\_52130}} & & \multicolumn{3}{c}{{415\_57112\_110099}} \\ \cline{2-4}\cline{6-8}\cline{10-12}\cline{14-16}
                               & $\mathrm{RPE}_{t}\downarrow$ & $\mathrm{RPE}_{r}\downarrow$ & ATE$\downarrow$ & & $\mathrm{RPE}_{t}$ & $\mathrm{RPE}_{r}$ & ATE & & $\mathrm{RPE}_{t}$ & $\mathrm{RPE}_{r}$ & ATE & & $\mathrm{RPE}_{t}$ & $\mathrm{RPE}_{r}$ & ATE \\
         \hline
         CF-3DGS    & 0.095 & 0.447 & 0.009   & & 0.140 & 0.401 & 0.021     & & 0.239 & 0.472 & 0.017     & & 0.110 & 0.424 & 0.014 \\
         SFGS       & 0.025 & 0.275 & 0.004   & & 0.093 & 0.331 & 0.020     & & 0.064 & 0.438 & 0.017     & & 0.049 & 0.351 & 0.024 \\
         Ours       & \cellcolor{gray!40}0.013 & \cellcolor{gray!40}0.080 & \cellcolor{gray!40}0.001    
                  & & \cellcolor{gray!40}0.012 & \cellcolor{gray!40}0.052 & \cellcolor{gray!40}0.001
                  & & \cellcolor{gray!40}0.005 & \cellcolor{gray!40}0.029 & \cellcolor{gray!40}0.001
                  & & \cellcolor{gray!40}0.004 & \cellcolor{gray!40}0.024 & \cellcolor{gray!40}0.001 \\
         \hline
    \end{tabular}
    }
    \caption{\textbf{Quantitative comparison of pose accuracy on CO3D-V2.} The unit of $\mathrm{RPE}_{r}$ is in degrees, ATE is in the ground truth scale and $\mathrm{RPE}_{t}$ is scaled by 100. 
    See the supplementary material for additional results.
    }
    \label{table:pose-T&T-co3dv2}
\end{table*}

\textbf{Implementation Details.}
Our implementation is primarily based on \emph{gsplat}~\cite{ye2024gsplatopensourcelibrarygaussian}, an accelerated 3DGS library. 
We modify the CUDA operator to backpropagate gradients for camera parameters.
All parameters are optimized using Adam.
For initialization, we optimize the relative pose between frames, and focal length. In joint process, the 3DGS parameters, absolute poses of cameras, and focal length are optimized. As vanilla 3DGS framework demonstrates inherent limitations in simultaneous optimization of all cameras within a single step. We introduce a gradient accumulation strategy~\cite{hermans2017accumulated} that enables co-optimization of the track Gaussians and all camera parameters in a bundle adjustment manner.
The camera pose is represented as a combination of an axis-angle representation $\mathrm{q}\in\mathfrak{so}(3)$ and a translation vector $\mathrm{t}\in\mathbb{R}^3$.
During training, we will clone new Gaussians from those links to the track points and apply the same training strategy as the original 3DGS (including clone, split, and delete). 
The number of track Gaussians is kept constant throughout training to ensure consistency with the imposed track constraints.
We set $\lambda=0.2$, $\lambda_{t2d}=0.01$, $\lambda_{t3d}=0.01$ in Eq.~\ref{eq:overall} for training.

\begin{figure}[!t]
  \centering
   \includegraphics[width=0.9\linewidth]{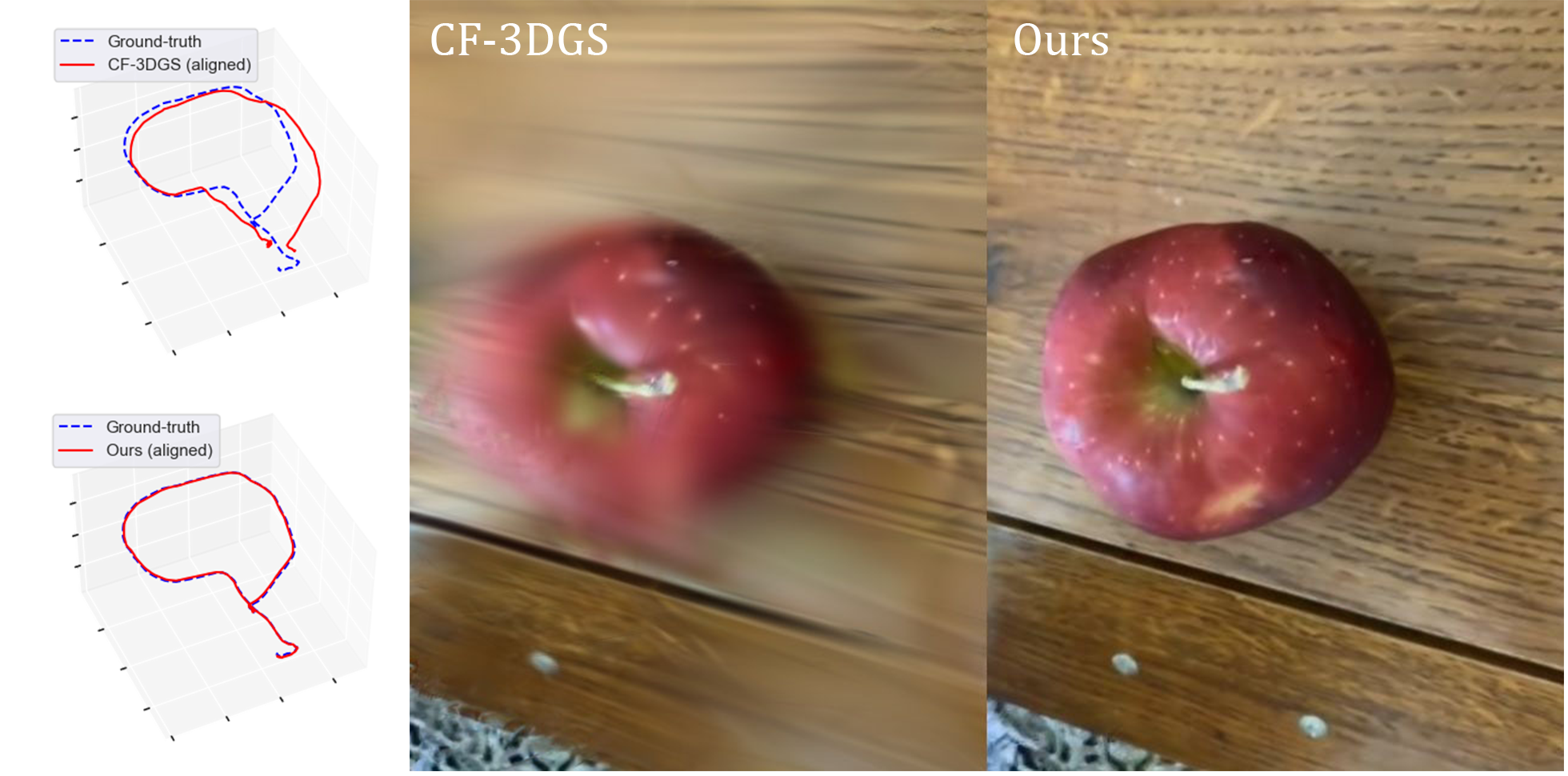}
   \caption{\textbf{Qualitative comparison for NVS and pose estimation on CO3D-V2.} Benefit from the accuracy of the camera pose estimation, the rendering quality of novel view synthesis obtained by our method is higher than CF-3DGS.}
   \label{fig:render-co3dv2}
\end{figure}

\subsection{Experimental Results and Analysis}

\textbf{Novel View Synthesis.}
Since the camera poses of test views are unknown, we need to estimate them for rendering. Following CF-3DGS~\cite{CF-3DGS-2024}, we obtain these test-view poses by minimizing the photometric error between the synthesized images and the test views using the pre-trained 3DGS model. We apply the same procedure to all baseline methods to maintain a consistent bias for a fair comparison.

We report the results on CO3D-V2 and Tanks and Temples in Tab.~\ref{table:render-T&T-co3dv2-syn}. Our method consistently outperforms all baselines on these real-world datasets with complex camera motions. 
Both CF-3DGS and SFGS are less effective on these challenging datasets. The main reason is that these methods rely heavily on local constraints. 
When the training sequences are sparsely sampled (e.g., 20 fps $\to$ 4 fps), photometric loss and relative pose estimation are highly prone to failure and become ineffective. However, our proposed track loss effectively imposes global geometric constraints under such conditions.
As illustrated in Fig.~\ref{fig:render-T&T} and~\ref{fig:render-co3dv2}, the advantages of our algorithm are well demonstrated, especially with large camera motions. Due to global joint optimization, multi-view geometric consistency is better maintained in the trained 3DGS model, leading to high-quality images. 

For further comparison, we evaluated our method on the Synthetic Dataset in Tab.~\ref{table:render-T&T-co3dv2-syn}, which features extremely complex camera motions. One result is shown in the bottom-right of Fig.~\ref{fig:teaser}. In this case, the camera not only moves in multiple circles around the object but also changes significantly in the vertical direction. Our synthesized image from the novel view remains clear and sharp, whereas the CF-3DGS result is blurry with obvious artifacts.

\begin{figure}[!t]
  \centering
  \includegraphics[width=\linewidth]{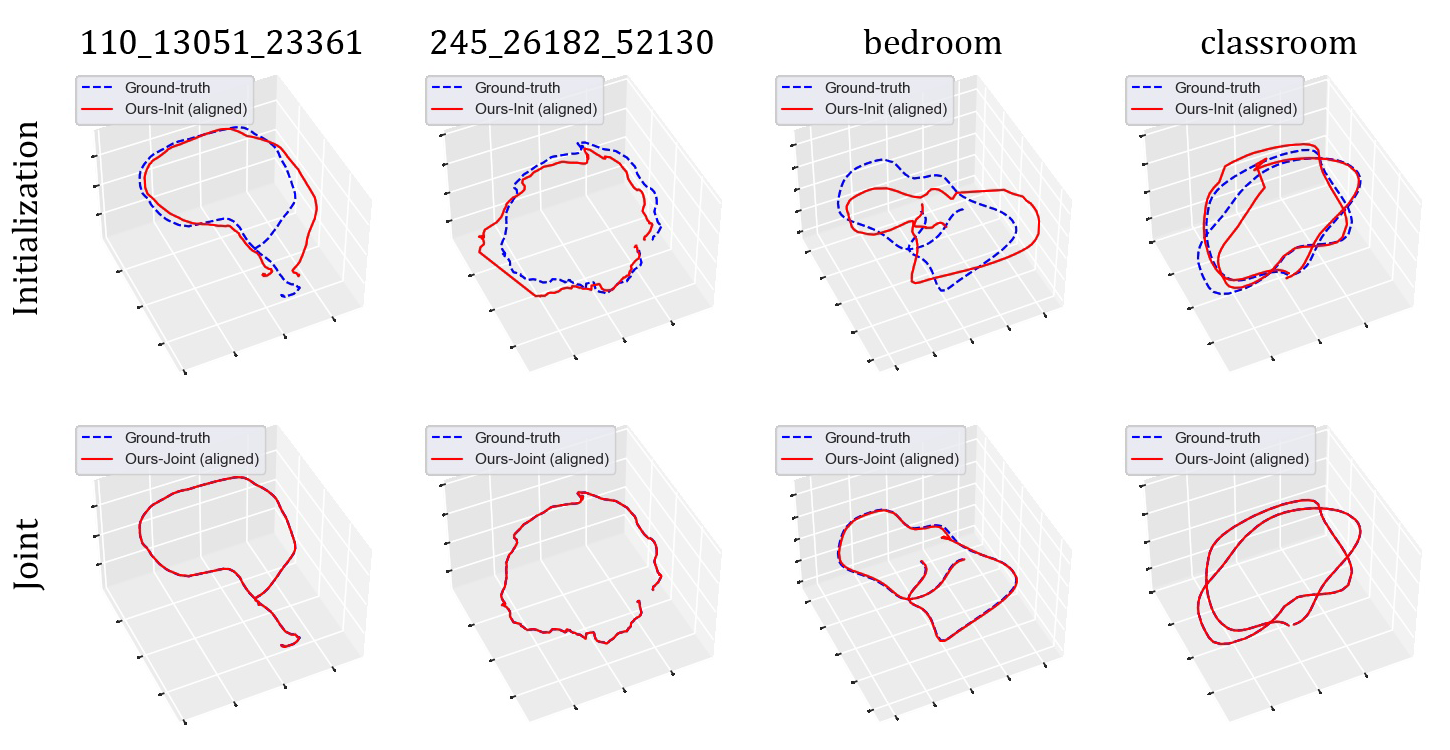}
   \caption{\textbf{The trajectory of initial stage and joint stage}. Our joint stage significantly improved the accuracy of camera pose.}
   \label{fig:render-phase}
\end{figure}

\begin{table*}[!t]
 \setlength{\tabcolsep}{1mm}
    \centering
    {\fontsize{8}{9}\selectfont
    \begin{tabular}{c |cccc |cccc |cccc |cccc} \hline
    \multirow{2}{*}{Scenes} & \multicolumn{4}{c|}{classroom} & \multicolumn{4}{c|}{{lego\_c2}} & \multicolumn{4}{c|}{livingroom} & \multicolumn{4}{c}{bedroom}\\ 
    & FoV($^\circ$) & $\mathrm{RPE}_{t}$ & $\mathrm{RPE}_{r}$ & ATE & FoV($^\circ$) & $\mathrm{RPE}_{t}$ & $\mathrm{RPE}_{r}$ & ATE & FoV($^\circ$) & $\mathrm{RPE}_{t}$ & $\mathrm{RPE}_{r}$ & ATE & FoV($^\circ$) & $\mathrm{RPE}_{t}$ & $\mathrm{RPE}_{r}$ & ATE \\ \hline
    CF-3DGS     & 0.993 & 0.588 & 2.436 & 0.07412    & 0.021 & 1.126 & 4.946 & 0.10795   & 0.029 & 0.425 & 2.104 & 0.07653  & 0.042 & 0.366 & 1.103 & 0.06284 \\
    SFGS        & 0.993 & 0.605 & 2.148 & 0.07811    & 0.021 & 1.101 & 4.519 & 0.10904   & 0.029 & 0.400 & 1.817 & 0.07356  & 0.042 & 0.343 & 1.075 & 0.06492 \\
    COLMAP      & 0.993 & 0.004 & 0.018 & 0.00023     & \cellcolor{gray!40}0.021 & 0.009 & 0.026 & 0.00019   & 0.029 & 0.008 & 0.026 & 0.00014   & 0.042 & \cellcolor{gray!40}0.009 & \cellcolor{gray!40}0.035 & \cellcolor{gray!40}0.00023 \\
    Ours        & \cellcolor{gray!40}0.012 & \cellcolor{gray!40}0.002 & \cellcolor{gray!40}0.013 & \cellcolor{gray!40}0.00008     & 0.031 & \cellcolor{gray!40}0.002 & \cellcolor{gray!40}0.015 & \cellcolor{gray!40}0.00011   & \cellcolor{gray!40}0.012 & \cellcolor{gray!40}0.002 & \cellcolor{gray!40}0.013 & \cellcolor{gray!40}0.00009   & \cellcolor{gray!40}0.003 & 0.013 & 0.062 & 0.00044 \\\hline
    \end{tabular}
    }
    \caption{\textbf{Quantitative comparison of parameter accuracy on our Synthetic Dataset.} We convert the estimated camera intrinsics focal to FoV and perform the errors of FoV with ground truth (provided by our synthetic datasets).
    As CF-3DGS and SFGS require the camera intrinsic parameters as fixed inputs, we set them the same as COLMAP+3DGS.}
    \label{table:pose-Synthetic}
\end{table*}

\begin{table}[!t]
\setlength{\tabcolsep}{1mm}
    \centering
    {\fontsize{8}{9}\selectfont
        \begin{tabular}{c l c c c c c}\hline
             ID & Variant       & PSNR$\uparrow$ & SSIM$\uparrow$  & LPIPS$\downarrow$ & ATE$\downarrow$ & FoV$\downarrow$ \\ \hline
             1  & COLMAP + 3DGS & 32.26          & 0.91            & 0.18              & 0.00020         & 0.271 \\ 
             2  & CF-3DGS       & 17.30          & 0.55            & 0.51              & 0.08036         & 0.271 \\ 
             3  & SFGS          & 14.48          & 0.50            & 0.59              & 0.08141         & 0.271 \\ \hline
             4  & w.o. 2D track & 18.18	        & 0.56            & 0.47              & 0.02020         & 2.617 \\
             5  & w.o. 3D track & 32.38          & 0.91            & 0.17              & 0.00275         & 0.063 \\
             6  & Ours          & \cellcolor{gray!40}32.58          & \cellcolor{gray!40}0.92            & \cellcolor{gray!40}0.16              & \cellcolor{gray!40}0.00018         & \cellcolor{gray!40}0.015 \\ \hline
        \end{tabular}
    }
    \caption{{Ablation study on our Synthetic Dataset.}}
    \label{tab:ablation-all}
\end{table}

\begin{table}[!t]
\setlength{\tabcolsep}{1mm}
    \centering
    {\fontsize{8}{9}\selectfont
    \begin{tabular}{c |cc |cc |cc} \hline
    \multirow{2}{*}{Scenes} & \multicolumn{2}{c|}{Ours} & \multicolumn{2}{c|}{Ours(FoV=60\degree)} & \multicolumn{2}{c}{COLMAP+3DGS}\\ 
                            & PSNR & LPIPS & PSNR  & LPIPS & PSNR  & LPIPS \\ \hline
    classroom       & \cellcolor{gray!40}36.26 & \cellcolor{gray!40}0.15    & 34.25 & 0.16   & 35.81 & 0.15 \\
    {lego\_c2}      & \cellcolor{gray!40}29.36 & \cellcolor{gray!40}0.12    & 23.82 & 0.26   & 28.77 & 0.15 \\
    livingroom      & \cellcolor{gray!40}33.52 & \cellcolor{gray!40}0.24    & 26.31 & 0.32   & 32.74 & 0.27 \\
    bedroom         & 31.17 & \cellcolor{gray!40}0.13    & 24.01 & 0.25   & \cellcolor{gray!40}31.73 & \cellcolor{gray!40}0.13 \\   \hline
    \end{tabular}
    }
    \caption{{NVS results on our Synthetic Dataset compared with Fixed FoV and COLMAP-assisted 3DGS.}
    }
    \label{tab:ablation-intrinsic}
\end{table}

\textbf{Camera Parameter Estimation.}
First, we compare the camera pose estimation with baseline methods. In the comparison, our method only assumes a constant camera focal length across frames, while others additionally input the camera focal length. The estimated camera poses are analyzed by Procrusts as in CF-3DGS and compared with the ground-truth of training views.
The quantitative results of camera pose estimation on CO3D-V2 datasets are summarized in Tab.~\ref{table:pose-T&T-co3dv2}.
The results show that our estimated camera parameters achieve the smallest error among all methods, with the Absolute Trajectory Error (ATE) being only one-tenth of that of the second-best method.
This demonstrates that our algorithm excels in scenes with complex camera motions. Compared to the baselines, the global tracking information we use eliminates accumulated errors, leading to more accurate camera pose estimation. Additionally, joint optimization enhances the stability of the estimation results. 

Next, we evaluate camera parameter estimation on our Synthetic Dataset. Tab.~\ref{table:pose-Synthetic} shows the estimation errors from CF-3DGS, SFGS, COLMAP, and ours. Note that CF-3DGS and SFGS use the camera FoV estimated by COLMAP. We find that our estimated camera FoVs and poses are comparable to those of COLMAP, and the camera pose error is 50 times smaller than baselines. Fig.~\ref{fig:render-phase} visualizes our estimated poses in different stages. Thanks to the joint optimization based on global track and the back-propagation of the gradient of the camera parameters, our approach is able to combine these two tasks, reducing the input requirements.

\subsection{Ablation Study}
\begin{figure}[!t]
  \centering
  \includegraphics[width=0.9\linewidth]{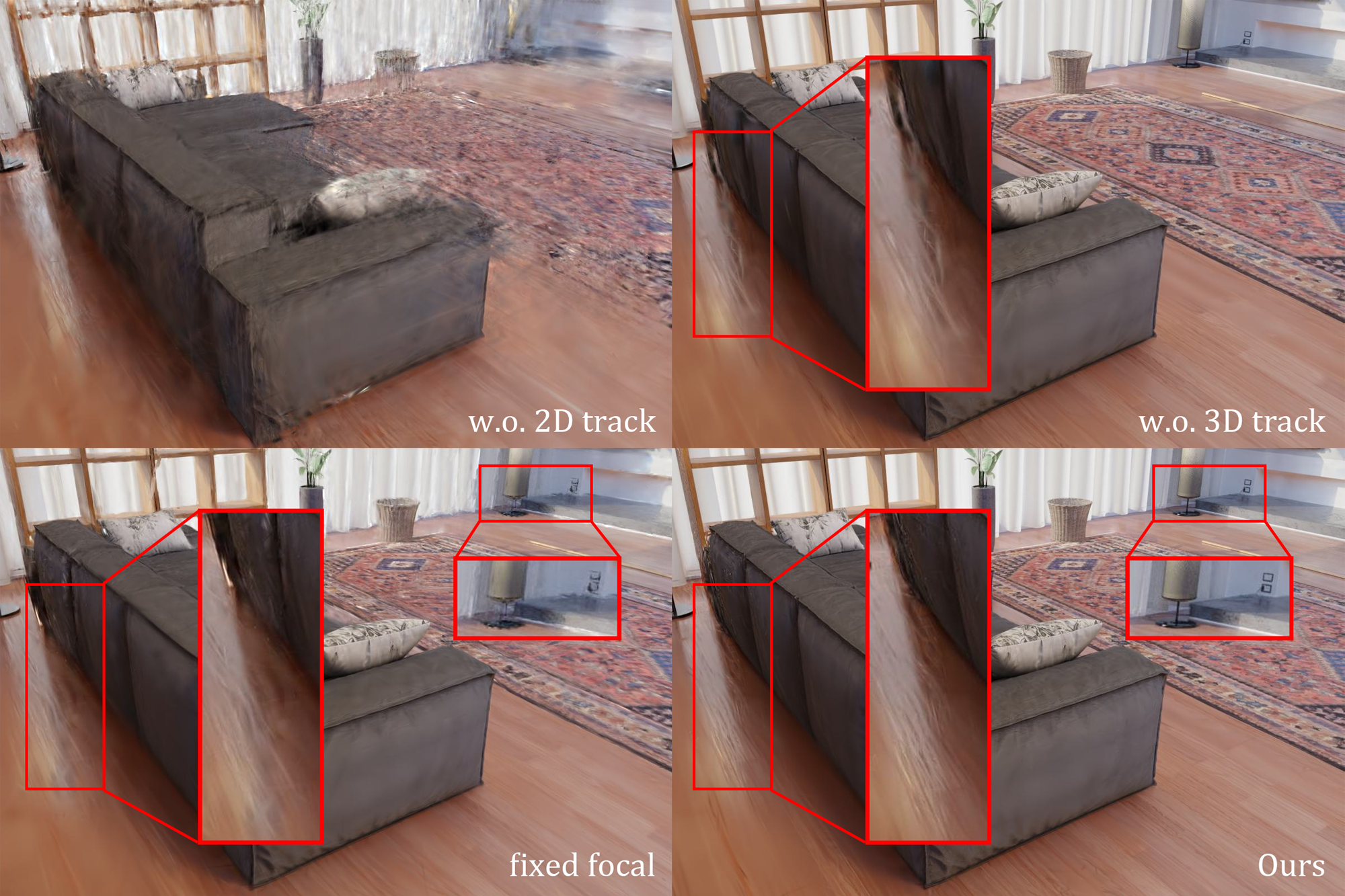}
   \caption{\textbf{Visualization of ablation study on Synthetic Dataset.} Without 2D/3D track loss or intrinsic optimization, the result appears blurry in novel view synthesis.}
   \label{fig:ablation-wotrack}
\end{figure}

\textbf{Effectiveness of Different Losses.}
We ablate each loss of the algorithm on Synthetic Dataset, since it has ground-truth camera parameters. Tab.~\ref{tab:ablation-all} reports the average synthesis quality and camera parameter errors across different algorithm variants (see supplementary material for details). 
First, any variant of our algorithm (Variant 4, 5, 6) is better than the baseline methods (Variant 2, 3) in synthesis quality and absolute camera position.
Second, Variant 4 shows that 2D track loss plays a crucial role in the entire joint optimization. When 2D track loss is not used, compared with the final method (Variant 4 vs. 6), there is a significant decrease in synthesis quality (18.18 vs. 32.58), and the camera parameter error is significantly larger (2.617 vs. 0.015). This shows that the reprojection error constrained by global consistency can significantly enhance the camera parameter estimation, thereby improving the 3DGS training effect and improving the new perspective synthesis ability, as illustrated in Fig. \ref{fig:ablation-wotrack}.
In addition, the results of Variant 5 vs. 6 show that 3D track loss can further enhance the geometric consistency of 3DGS. Using 3D track loss, the PSNR of NVS can be further improved by 0.2 dB, and the ATE of the camera can be reduced by an order of magnitude.

\textbf{Effectiveness of Intrinsic Optimization.}
Accurate camera intrinsics resolve scale ambiguity in 3DGS models, leading to improved NVS performance. As shown in Tab.~\ref{table:pose-Synthetic}, our method produces more accurate intrinsics (i.e., FoV) compared to COLMAP with vanilla 3DGS. We also performed an ablation study with a fixed camera FoV of $60^\circ$ and without further optimization. The results, shown in Tab.~\ref{tab:ablation-intrinsic}, indicate a 21.4\% average decrease in PSNR, due to scale ambiguity introduced by inaccurate camera intrinsics.

\textbf{Comparison with COLMAP-Assisted 3DGS.}
We compare the NVS results generated by our method against vanilla 3DGS, where the camera intrinsics and extrinsics are estimated using COLMAP on Synthetic Dataset. 
Tab.~\ref{tab:ablation-intrinsic} shows that our method outperforms COLMAP-assisted 3DGS in most scenes. 
Unlike the original 3DGS, which uses a fixed camera pose for training, our method seamlessly integrates 3DGS training with camera parameter estimation, allowing the two tasks to complement each other and ultimately achieve high-quality novel view synthesis.

\section{Conclusion}
\label{sec:conclution&limitation}
We presented TrackGS, a novel framework that successfully integrates global feature tracks with 3D Gaussian Splatting to achieve COLMAP-free novel view synthesis. By establishing geometric constraints through track Gaussians and introducing novel 2D/3D consistency losses, our method jointly optimizes camera parameters and scene representation with high accuracy. The differentiable formulation of camera intrinsics completes the pipeline, removing all dependencies on precomputed parameters. Experiments demonstrate significant improvements over state-of-the-art methods in both rendering quality and pose estimation accuracy across diverse scenarios.
Future extensions include extending the framework to per-view intrinsics, distortion modeling, and dynamic scene support.

\section*{Acknowledgements}
This work was partly supported by the Anhui Provincial Natural Science Foundation (2508085MA001).

\bibliography{aaai2026}

\end{document}


\maketitle

\section{Derivation of Camera Intrinsic Parameters}
\label{sec:focal-div}
\begin{figure}[t]
  \centering
   \includegraphics[width=0.8\linewidth]{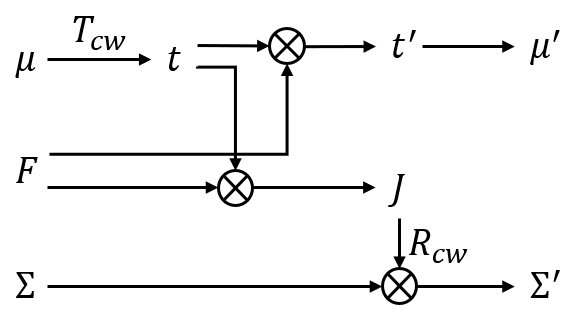}
   \caption{\textbf{Computational graph of parameters}.}
   \label{fig:focal-div}
\end{figure}

With the output image width and height $(w,h)$, as well as the near and far clipping planes $(n,f)$, the extrinsic matrix $T_{cw}$ and the projection matrix $P$, representing the transformation from camera space to normalized clip space, are denoted as follows:
\begin{equation}\label{eq:supp-T&P}
    T_{cw}=\begin{bmatrix}R_{cw}&t_{cw}\\0&1\end{bmatrix}, \\
    P=\begin{bmatrix}
        \frac{2f_{x}}{w} & 0 & 0 & 0 \\
        0 & \frac{2f_{y}}{h} & 0 & 0 \\
        0 & 0 & \frac{f+n}{f-n} & \frac{-2fn}{f-n} \\
        0 & 0 & 1 & 0 \\
    \end{bmatrix}.
\end{equation}
Fig.~\ref{fig:focal-div} demonstrates the computational graph of parameters, which are related to camera intrinsics. In our discussion, the focal length is $F=(f_{x},f_{y})$ and the principal point is $(c_{x},c_{y})$.
For a 3D Gaussian parameterized by its mean $\mu\in\mathbb{R}^3$ and covariance $\Sigma\in\mathbb{R}^{3\times 3}$, the loss $\mathcal{L}$ is formulated by its 2D projected mean $\mu^{'}$ and covariance $\sigma^{'}$.
We convert the mean $\mu$ into $t=(t_{x},t_{y},t_{z},t_{w})\in \mathbb{R}^4$ in camera coordinates, $t^{'}=(t_{x}^{'},t_{y}^{'},t_{z}^{'},t_{w}^{'})\in\mathbb{R}^4$ in normalized coordinates (NDC),
and finally $\mu^{'}\in\mathbb{R}^{2}$ in pixel coordinates as follows:
\begin{equation}\label{eq:supp-t&mu}
    t=T_{cw}\begin{bmatrix}\mu & 1\end{bmatrix}^\top,
    t^{'}=Pt,
    \mu^{'}=\begin{bmatrix}
        \frac{1}{2}(\frac{w\cdot t_{x}^{'}}{t_{w}^{'}}+1)+c_{x} \\
        \frac{1}{2}(\frac{h\cdot t_{y}^{'}}{t_{w}^{'}}+1)+c_{y}
    \end{bmatrix}.
\end{equation}
Notice that the projection of a 3D Gaussian does not result in a 2D Gaussian, the projection of $\Sigma$ to pixel coordinates is approximated with a first-order Taylor expansion at $t$ in camera space, then the affine transform $J\in\mathbb{R}^{2\times 3}$ and the 2D covariance $\Sigma^{'}\in\mathbb{R}^{2\times 2}$ are:
\begin{equation}\label{eq:supp-J&sigma}
    J=\begin{bmatrix}
        \frac{f_{x}}{t_{z}} & 0 & -\frac{f_{x}\cdot t_{x}}{t_{z}^{2}}\\
        0 & \frac{f_{y}}{t_{z}} & -\frac{f_{y}\cdot t_{y}}{t_{z}^{2}}
    \end{bmatrix},
    \Sigma^{'}=J R_{cw} \Sigma R_{cw}^\top J^\top.
\end{equation}
Given the gradients of $\mathcal{L}$ with respect to 2D mean $\mu^{'}$ and covariance $\Sigma^{'}$, we can back-propagate the gradient of focal length $F$ as:
\begin{equation}\label{eq:supp-gradF}
    \frac{\partial \mathcal{L}}{\partial F}=\frac{\partial \mathcal{L}}{\partial \mu^{'}} \frac{\partial \mu^{'}}{\partial F} + \frac{\partial \mathcal{L}}{\partial \Sigma^{'}}\frac{\partial \Sigma^{'}}{\partial F}.
\end{equation}

First, we compute the gradient contribution of 2D mean $\mu^{'}$ to focal length $F$, $\frac{\partial \mu^{'}}{\partial F}$ can be obtained by the chain rule:
\begin{equation}\label{eq:supp-gradF-mu}
\begin{aligned}
    \frac{\partial \mathcal{\mu^{'}}}{\partial F} & =  \frac{\partial \mu^{'}}{\partial t^{'}} \frac{\partial t^{'}}{\partial F}\\
     &=\begin{bmatrix}
         \frac{w}{2 t_{w}^{'}} & 0 & 0 & -\frac{w t_{x}^{'}}{(t_{w}^{'})^{2}} \\
        0 & \frac{h}{2 t_{w}^{'}} & 0 & -\frac{h t_{y}^{'}}{(t_{w}^{'})^{2}}
     \end{bmatrix}
     \begin{bmatrix}
         \frac{2 t_{x}}{w} & 0 & 0 & 0 \\ 0 & \frac{2 t_{y}}{h} & 0 & 0
     \end{bmatrix}^\top\\
     &=\begin{bmatrix}
         \frac{t_{x}}{t_{z}} & 0 \\
         0 & \frac{t_{y}}{t_{z}}
     \end{bmatrix},
\end{aligned}
\end{equation}
where $t_{w}^{'}=t_{z}$ from Eq.~\ref{eq:supp-t&mu}.

Then, for the second part of Eq.~\ref{eq:supp-gradF}, we use another parameter $J$ to compute this component, which means 
$\frac{\partial \mathcal{L}}{\partial \Sigma^{'}}\frac{\partial \Sigma^{'}}{\partial F}=\frac{\partial \mathcal{L}}{\partial J}\frac{\partial J}{\partial F}$.
\textit{gsplat}~\cite{ye2024gsplatopensourcelibrarygaussian} obtained the gradient of $\mathcal{L}$ to the affine transform $J$ through $T=J R_{cw}\in \mathbb{R}^{2\times 3}$ as:
\begin{equation}\label{eq:supp-gradJ-L}
    \partial \mathcal{L}=<\frac{\partial \mathcal{L}}{\partial T}{R_{cw}^\top},\partial J>,\mathrm{where}\frac{\partial \mathcal{L}}{\partial T}=\frac{\partial \mathcal{L}}{\partial \Sigma^{'}}T\Sigma^\top+{\frac{\partial \mathcal{L}}{\partial \Sigma^{'}}}^\top T\Sigma,
\end{equation}
with the gradient of $J$ to the focal length $F$ as:
\begin{equation}\label{eq:supp-gradF-J}
    \frac{\partial J}{\partial f_{x}}=\begin{bmatrix}
        \frac{1}{t_{z}} & 0 & -\frac{t_{x}}{t_{z}^{2}} \\
        0 & 0 & 0
    \end{bmatrix},
    \frac{\partial J}{\partial f_{y}}=\begin{bmatrix}
        0 & 0 & 0 \\
        0 & \frac{1}{t_{z}} & -\frac{t_{y}}{t_{z}^{2}}
    \end{bmatrix}.
\end{equation}

\begin{figure*}[!t]
  \centering
   \includegraphics[width=\linewidth]{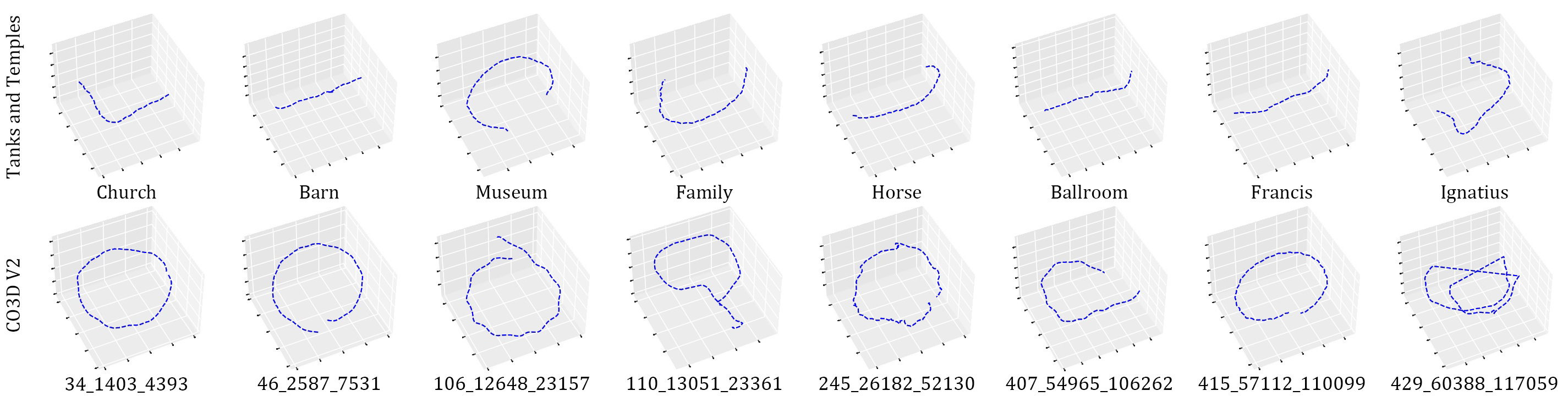}
   \caption{\textbf{Ground-truth camera trajectory for Tanks and Temples and CO3D-V2}.}
   \label{fig:suppl-gt-poses}
\end{figure*}

Finally, the gradient of loss $\mathcal{L}$ with respect to focal length $F$ in Eq.~\ref{eq:supp-gradF} is formulated as:
\begin{equation}
    \left\{
    \begin{aligned}
        & \frac{\partial \mathcal{L}}{\partial f_{x}} = \frac{t_x}{t_z}\frac{\partial \mathcal{L}}{\partial \mu_{x}^{'}} + <\frac{\partial \mathcal{L}}{\partial T}{R_{cw}^\top},\frac{\partial J}{\partial f_{x}}>, \\
        & \frac{\partial \mathcal{L}}{\partial f_{y}} = \frac{t_y}{t_z}\frac{\partial \mathcal{L}}{\partial \mu_{y}^{'}} + <\frac{\partial \mathcal{L}}{\partial T}{R_{cw}^\top},\frac{\partial J}{\partial f_{y}}>. \\
    \end{aligned}
    \right.
\end{equation}

\section{Implementation Details}
We provide more details about the datasets and training procedure in following sections.
\subsection{Dataset}
We select three datasets for training and evaluation including existing datasets Tanks and Temples, CO3D-V2, and a virtual Synthetic dataset created by ourself.
Tab.~\ref{table:suppl-dataset-details} shows the details of the scenes in all three datasets, where $Max.\ rot$ is the maximum relative rotation angle between any two frames and the $Avg.\ adj.\ rot$ donates the average relative rotation angle between two adjacent frames. The later represents the magnitude of the relative angle change in a sequence. In comparison, notice that the frame changes in Tanks and Temples are quite small, and our Synthetic datasets are more complex than CO3D-V2, although the $Max.\ rot$ of both datasets is 180 degrees. The visualization of these camera trajectories is shown in Fig.~\ref{fig:suppl-gt-poses}.

\begin{figure*}[!t]
  \centering
  \includegraphics[width=0.66\linewidth]{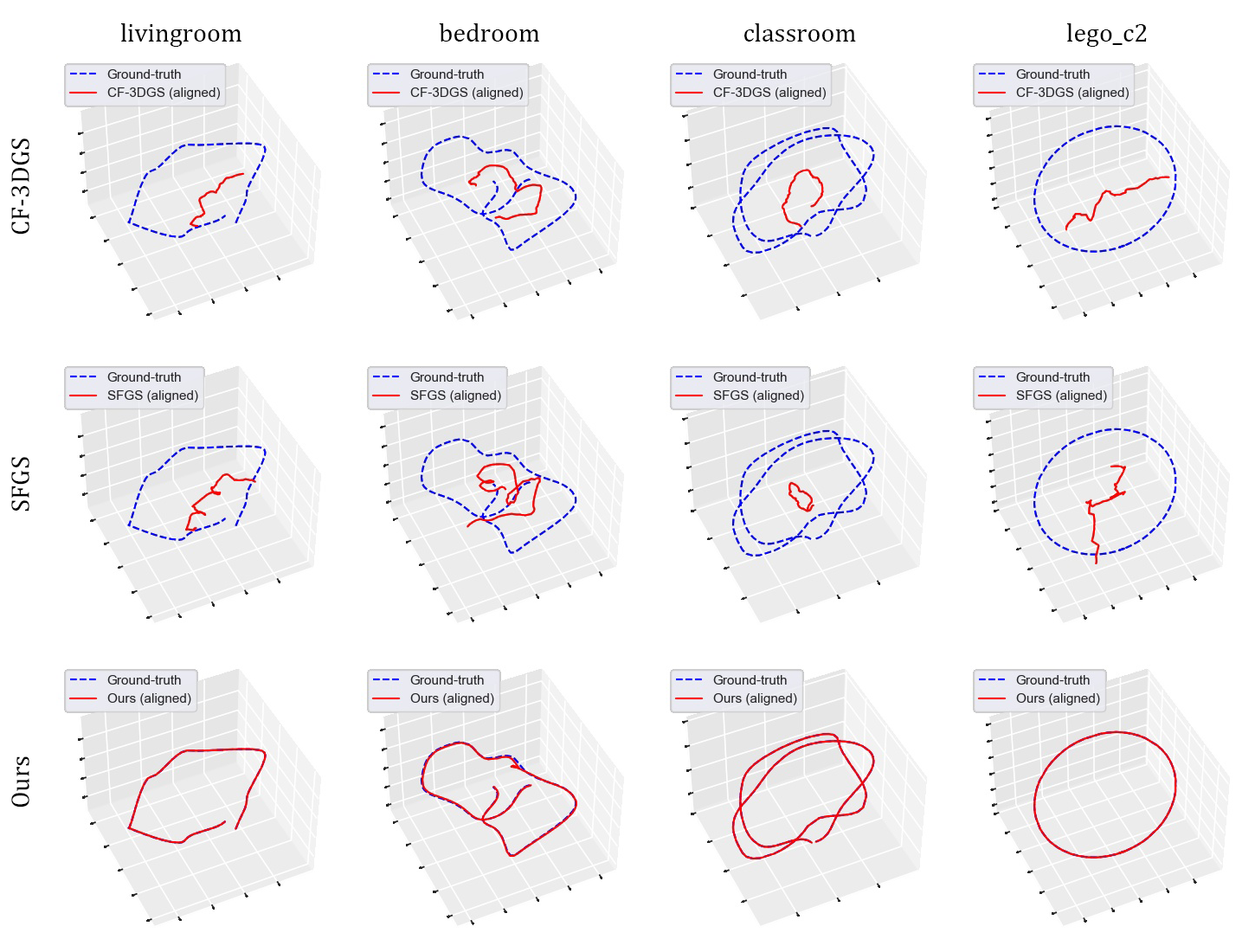}
   \caption{\textbf{Comparison of pose estimation on Synthetic dataset.}}
   \label{fig:suppl-pose-synthetic}
\end{figure*}

\textbf{Tanks and Temples.}
\red{The raw dataset comprises a set of indoor and outdoor scenes characterized by complex and long-range camera trajectories. CF-3DGS adopts a subset of 8 scenes from this dataset for evaluation, while their selected frame sequences are overly smooth and lack the rich camera dynamics typically observed in real-world scenarios. To better assess the robustness and generalization of our method under more realistic conditions, we re-sample the same 8 scenes from the original video sequences with a lower frame rate of 4 FPS. Moreover, we extend the sampling range to cover longer camera trajectories, thereby introducing more significant and large camera motions. This results in a more challenging benchmark that better reflects practical deployment conditions. A visualization of the re-sampled camera trajectories is provided in Fig. \ref{fig:suppl-pose-synthetic}.}

\textbf{Synthetic Dataset.}
Extracting accurate camera parameters from image sequences is challenging. While existing datasets employ COLMAP to derive these parameters, inaccuracies remain. To facilitate more precise comparisons, we have created a Synthetic Dataset using Blender~\cite{blender2018}. This dataset includes four indoor scenes. The camera movements in the \textit{classroom} (180 frames) and \textit{lego\_c2} (75 frames) scenes are object-centric, whereas those in the \textit{livingroom} (150 frames) and \textit{bedroom} (180 frames) scenes involve roaming. In both types of scenes, the camera navigates complex paths extending over 360 degrees. The visualization of the camera trajectories is illustrated in Fig. \ref{fig:suppl-pose-synthetic}.

\begin{table}[!ht]
    \centering
    \resizebox{\linewidth}{!}{
    \begin{tabular}{c c ccccc} \hline
        & {Scenes}                  & {Type}        & \makecell{Seq.\\length}      & \makecell{Frame\\ rate}      & \makecell{Max.\\ rot (deg)}      & \makecell{Avg. adj. \\rot (deg)}\\ \hline
    \multirow{8}{*}{\rotatebox{90}{Tanks and Temples}}    
        & Church                    &indoor         &152                &4                  & 173.2 & 0.98      \\
        & Barn                      &outdoor        &152                &4                  & 171.0 & 0.88       \\
        & Museum                    &indoor         &152                &4                  & 180.0 & 2.18      \\
        & Family                    &outdoor        &152                &4                  & 180.0 & 1.36      \\
        & Horse                     &outdoor        &152                &4                  & 180.0 & 1.41      \\
        & Ballroom                  &indoor         &152                &4                  & 162.0 & 1.17      \\
        & Francis                   &outdoor        &152                &4                  & 176.7 & 0.95      \\
        & Ignatius                  &outdoor        &152                &4                  & 180.0 & 1.54      \\ \hline
    \multirow{8}{*}{\rotatebox{90}{CO3D-V2}}  
        & {34\_1403\_4393}          &indoor         &202                &30                 &180.0      &1.53   \\
        & {46\_2587\_7531}          &indoor         &202                &30                 &180.0      &1.60   \\
        & {106\_12648\_23157}       &outdoor        &202                &30                 &180.0      &1.33   \\
        & {110\_13051\_23361}       &indoor         &202                &30                 &71.6       &0.99   \\
        & {245\_26182\_52130}       &indoor         &202                &30                 &180.0      &1.40   \\
        & {407\_54965\_106262}      &indoor         &202                &30                 &180.0      &1.46   \\
        & {415\_57112\_110099}      &outdoor        &202                &30                 &180.0      &1.60   \\
        & {429\_60388\_117059}      &outdoor        &202                &30                 &180.0      &3.11   \\ \hline
    \multirow{4}{*}{\rotatebox{90}{Synthetic}}
        & classroom                 &indoor         &180                &0                  &180.0      &3.84   \\
        & {lego\_c2}                &indoor         &75                 &0                  &180.0      &3.77   \\
        & livingroom                &indoor         &150                &0                  &180.0      &2.63   \\
        & bedroom                   &indoor         &180                &0                  &180.0      &2.33   \\ \hline
    \end{tabular}
    }
    \caption{\textbf{Details of the selected sequences.} }
    \label{table:suppl-dataset-details}
\end{table}

\begin{algorithm}
\caption{Initialization}\label{alg:suppl-init}
\begin{algorithmic}
\State $\mathcal{I}=\{I_{i}\}_{i=1}^{M} \gets$ {Input images}
\State DPT $\gets$ {Monocular Depth Estimation Model}
\State $\{p_{i}\} \gets \mathrm{SuperPoint}(\mathcal{I})$ \Comment{extract feature}
\State $(p_{i},p_{j})\gets \mathrm{SuperGlue}(\{p_{i}\})$ \Comment{image matching}
\State MST $\gets$ Kruskal Algorithm    \Comment{construct MST}
\State $\mathcal{P}=(P,\{p_{i}\}_{i=1}^{l})\gets \mathrm{Track(MST)}$ 
\State \Comment{extract track points}
\Statex\hrule
\Loop   \Comment{Loop 100 iterations}
    \For{edge $(i,j)$ in MST}
        \State $L_{reproj}(i,j)=\|K \cdot (T_{ji} \cdot K^{-1} \cdot p_{i}) - p_{j}\|$ 
        \State $L_{reproj}+=L_{reproj}(i,j)$
    \EndFor
    \State $K,\{T_{ji}\}\gets \min L_{reproj}$ 
\EndLoop\Comment{Optimize $T_{ji}, K$}
\For {$(P,\{p_{i}\})$ in $\mathcal{P}$}
    \State $d_{i} \gets \mathrm{DPT}(p_{i})$
    \State $P \gets \mathrm{InitTrackPoint}(K,\{d_i\},\{p_{i}\})$ 

\EndFor \Comment{Init Track Points}
\State $T_{cw}=\prod_{(i,j)}T_{ji}$ \Comment{Init camera extrinsic}
\Statex
\hrule
\State \textbf{return} $K,T_{cw},P$
\end{algorithmic}
\end{algorithm}

\begin{algorithm}
\caption{Joint optimization}\label{alg:suppl-joint}
\begin{algorithmic}
\State $lr\gets lr_{\mu},lr_q,lr_s,lr_{\alpha}$ \Comment{Initial lr of 3D Gaussians}
\State $lr_{pos}\gets lr_{R_{cw}},lr_{T_{cw}}$ \Comment{Initial lr of camera pose}
\State $G_{track}\gets \mathrm{Gaussian}(P)$ \Comment{Initial track 3D Gaussians}
\Statex
\hrule
\Procedure{WarmUp:}{}
    \State $\mathrm{step}=0$
    \While{$\mathrm{step < 500}$}
        \State $\hat{I}_i,\hat{p}_i \gets \mathrm{Rasterize}(G_{track},T_{cw,i},K)$
        \State $L_i \gets \mathcal{L}(I_i,p_i,\hat{I}_i,\hat{p}_i)$
        \State $L \gets \sum_i^M{L_i}$ 
        \State $G_{track},T_{cw},K \gets \mathrm{Adam}(\nabla{L})$ 
        \State \Comment{Update track Gaussians and camera param}
        \State $lr \gets \mathrm{schedule}(lr)$ \Comment{Update Gaussian lr}
        \State $lr_{pos} \gets \mathrm{schedule}(lr_{pos})$ \Comment{Update pose lr}
        \If{$\mathrm{step}<100$}
            \State $lr_{focal}=0.0$
        \Else
            \State $lr_{focal}=\max(1e^{-4},5e^{-3}*(1.0-\frac{\mathrm{step}}{500}))$
        \EndIf \Comment{Update focal lr}
        \State $\mathrm{step} \gets \mathrm{step} + 1$
    \EndWhile
\EndProcedure
\Statex
\hrule
\Procedure{Joint 3DGS:}{}
\State $\mathrm{step}=0$
\State $\tilde{\mathcal{I}} \gets \mathrm{Queue(shuffle}(\mathcal{I}))$ \Comment{Shuffle images}
\While{$\mathrm{step}<30000$}
    \State $I_i \gets \mathrm{QuePopLeft}(\tilde{\mathcal{I}})$ \Comment{Pop first elem in Que}
    \State $G \gets G_{track}+G_{normal}$
    \State $\hat{I}_i,\hat{p}_i \gets \mathrm{Rasterize}(G,T_{cw,i},K)$
    \State $L_i\gets \mathcal{L}(I_i,p_i,\hat{I}_i,\hat{p}_i)$
    \State $G\gets \mathrm{Adam}(\nabla{L_i})$ \Comment{Update Gaussians}
    \State $lr$ $\leftarrow$ $schedule(lr)$ \Comment{Update Gaussians lr}
    \If{$\tilde{\mathcal{I}}==\emptyset$}
        \State $L\gets \sum_i^M{L_i}$ 
        \State $T_{cw},K \gets \mathrm{Adam}(\nabla{L})$ 
        \State \Comment{Update all cameras param}
        \State $lr_{pos},lr_{focal}\gets \mathrm{schedule}(lr_{pos},lr_{focal})$ 
        \State \Comment{Update all cameras lr}
        \State $\tilde{\mathcal{I}} \gets \mathrm{Queue(shuffle}(\mathcal{I}))$ \Comment{Shuffle images}
    \EndIf
    \For{all $(\mu,\sum,c,\alpha)$ in G}
        \If{$\nabla_{p}L<\tau_p$} \Comment{Densification}
            \State {$\mathrm{SplitGaussians}(\mu,\sum,c,\alpha)$}
            \State {$\mathrm{CloneGaussians}(\mu,\sum,c,\alpha)$}
        \EndIf
        \If{$\alpha < \epsilon$ or IsTooLarge$(\mu,\sum)$}
            \State $\mathrm{RemoveGaussian}(G_{normal})$ \Comment{Pruning}
        \EndIf
    \EndFor \Comment{Adaptive control of Gaussians}
    \State $\mathrm{step} \gets \mathrm{step} + 1$
\EndWhile
\EndProcedure
\end{algorithmic}
\end{algorithm}

\subsection{Training Details}

\textbf{Initialization.}
To provide the initial values of camera parameters and 3D Gaussians, we first obtain mono-depth maps of images $\mathcal{I}$ by DPT~\cite{dpt2021iccv}. We extract the feature points $\{p_{i}\}$ of each image $I$ with SuperPoint~\cite{detone2018superpoint} and compute the feature matches among all images with SuperGlue~\cite{sarlin20superglue}. Then we construct the Maximum Spanning Tree (MST) by Kruskal’s algorithm, where the node represents each image and the weight of each edge is determined by the number of feature matching pairs between two images. We extract the set of tracks $\mathcal{P}$ from the MST, where each element $(P,\{p_{i}\})\in \mathcal{P}$ is a 3D track point $P$ and its corresponding matching points $\{p_{i}\}$ associated with the training images. Later on, we define a reprojection loss for the edges of MST, and minimize this objective to optimize the camera intrinsic $K$ and the relative camera extrinsic matrix $T_{ji}$. This optimization procedure is set as 100 steps in our experiment. Finally, we obtain the initialization of the location of 3D track points $P$, the camera intrinsic $K$ and extrinsic matrix $T_{cw}$. The whole algorithm of initialization is summarized in Alg. \ref{alg:suppl-init}.

\begin{table}[h]
\label{tab:lr}
\centering
\resizebox{\columnwidth}{!}{
\begin{tabular}{c|c|c|c|c|c}
\hline
$lr_{\mu}$ & $lr_q$ & $lr_s$ & $lr_{\alpha}$ & $lr_{R_{cw}}$ & $lr_{T_{cw}}$ \\ \hline
$1.6 * xyz\_scale * 1e^{-2}$ & $1e^{-3}$ & $5e^{-3}$ & $5e^{-2}$ & $5e^{-3}$ & $1e^{-2}$ \\ \hline
\end{tabular}
}
\end{table}
\textbf{Joint optimization.}
As detailed in Alg.~\ref{alg:suppl-joint}, since the initial camera poses and tracking points are very noisy, before training of novel view synthesis, we take a warmup to get more precise parameters. 
Here, we draw on the concept of global bundle adjustment, first optimizing these parameters through RGB loss and 2D/3D track loss over 500 epochs. The initial learning rate for each variable is set as above.

For the learning rate of $\mu$, considering the global scale of scene, we introduce the bounding sphere radius of the initial point clouds $xyz\_scale$ as a parameter. Additionally, the learning rates for both $\mu$ and the camera parameters are decayed using the \textit{ExponentialLR} mechanism. The remaining learning rates are kept constant.
Moreover, the leanring rate for the focal length is set differentially: it is set to 0 during the initial 100 epochs, meaning that only the camera's pose and the initial 3DGS will be optimized. After the first 100 epochs, it decreases according to the following formulation:
\begin{equation}
\resizebox{\columnwidth}{!}{%
    $lr_{focal} = \left\{
\begin{array}{rl}
0.0 &, step \leq 100 \\
\max(1e^{-4},5e^{-3}*(1.0-step/500)) &, step > 100
\end{array} \right.$%
}
\end{equation}
During warmup, our goal is to achieve a better geometric initialization, at which point the weights of the RGB loss and 2D/3D track loss are all set to $1.0$. 

During the warmup, we do not perform clone, split, and prune operations on the Gaussian kernels to ensure better geometric constraints on the track points. Afterwards, we will clone new Gaussian kernels from those associated with the track points and apply the same training strategy as the original 3DGS (including clone, split, and prune). However, the tracked Gaussians still need to be preserved without pruning. The learning rate of $\mu$ and camera parameters continue to decay using \textit{ExponentialLR} from the end of the warmup and the other learning rates still remain constant. For optimization of the camera parameters,we update the camera parameters after calculating the loss of all cameras like bundle adjustment (BA). Considering the limitations of GPU resources, we optimize the Gaussians separately for each camera.

\section{Additional Experiments and Results}
We present additional results of novel view synthesis and camera parameter estimation by our method and other baselines, including SFGS~\cite{ji2024sfmfree3dgaussiansplatting} and CF-3DGS~\cite{CF-3DGS-2024} 
, on Tanks and Temples, CO3D-V2, and Synthetic Dataset.

\subsection{Novel View Synthesis}
\begin{figure*}[!t]
  \centering
  \includegraphics[width=\linewidth]{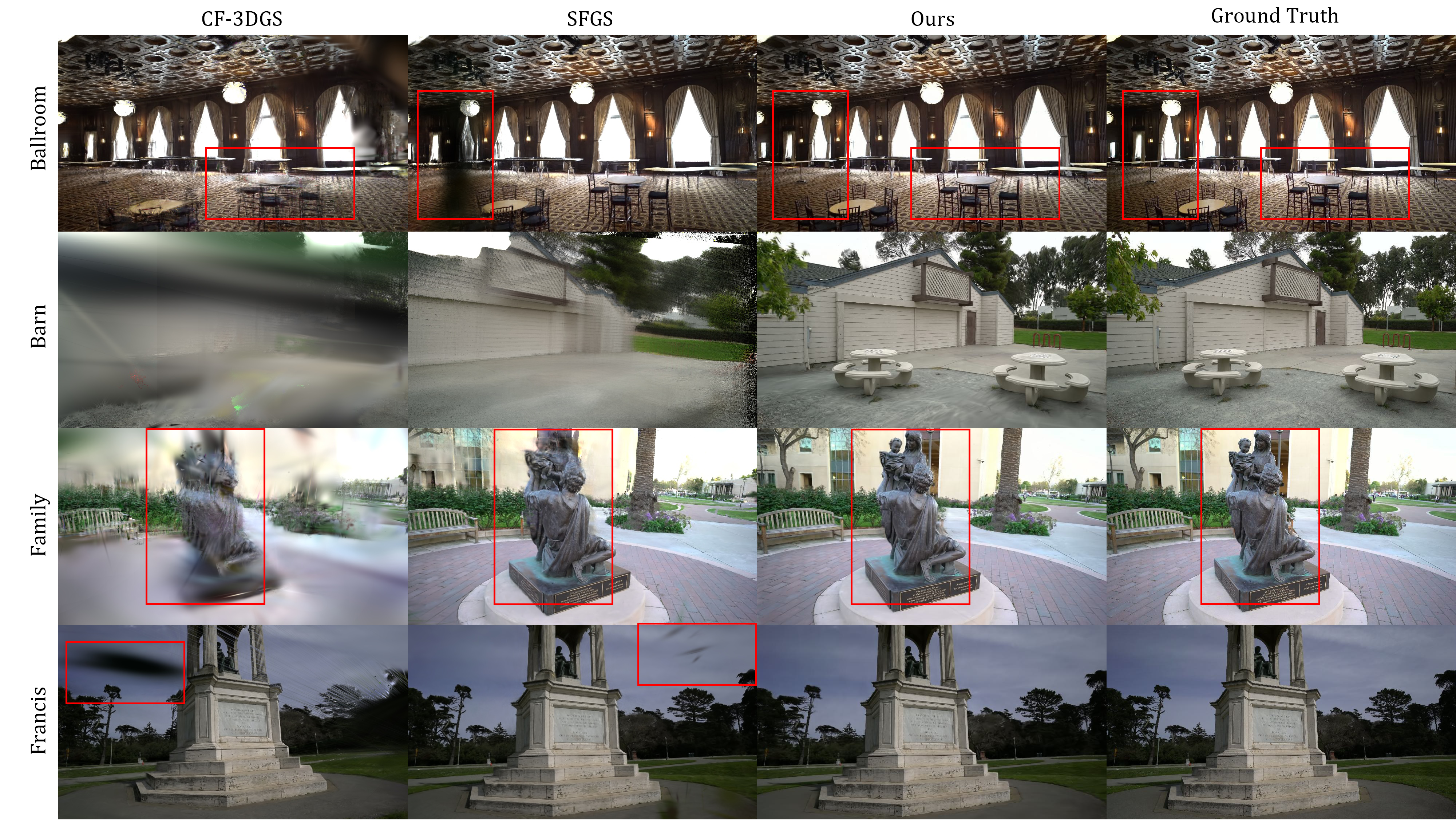}
   \caption{\textbf{Comparison of novel view synthesis results on Tanks and Temples.}}
   \label{fig:suppl-render-T&T}
\end{figure*}

\begin{figure*}[!t]
  \centering
  \includegraphics[width=\linewidth]{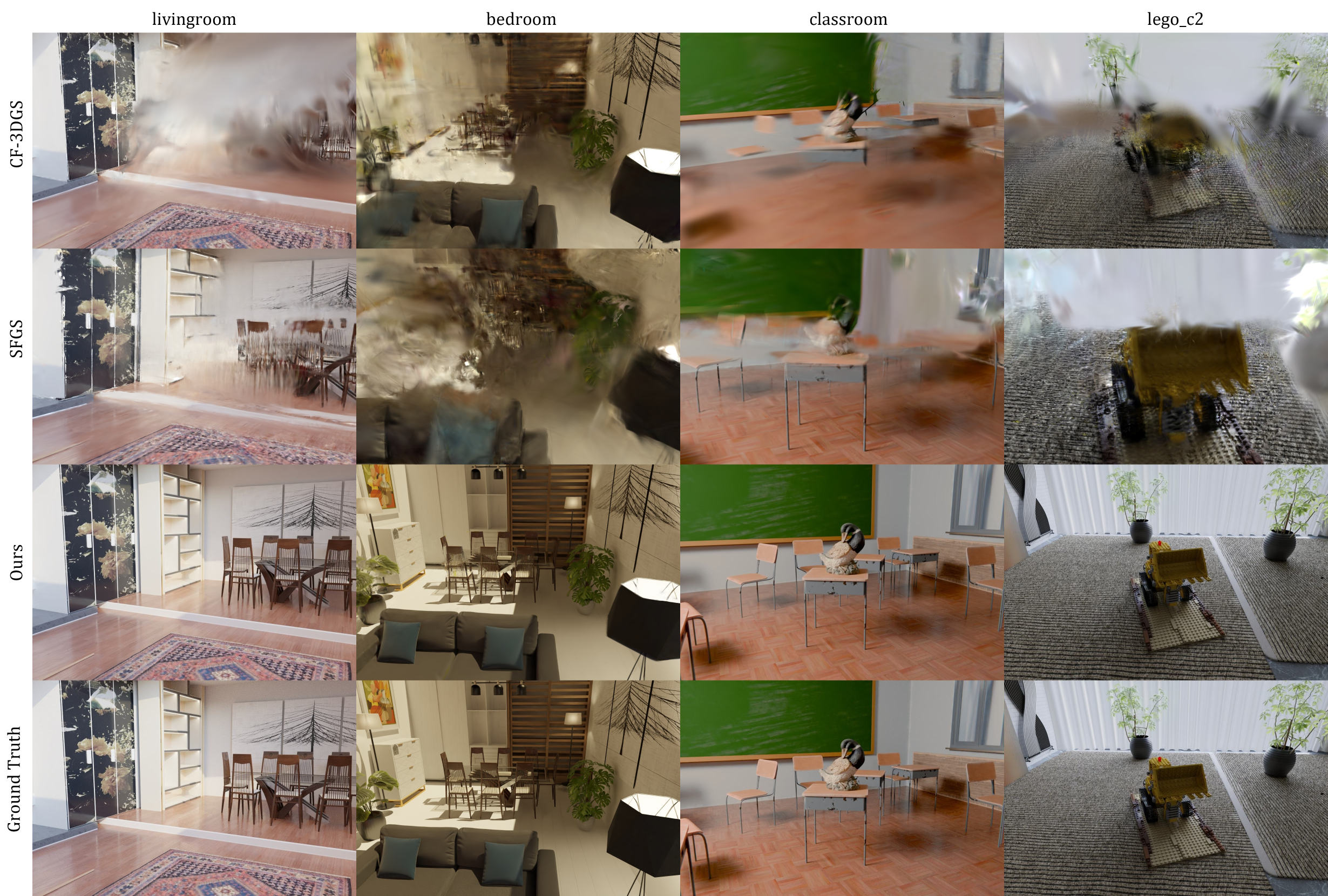}
   \caption{\textbf{Comparison of novel view synthesis results on Synthetic dataset.}}
   \label{fig:suppl-render-synthetic}
\end{figure*}

\begin{figure*}[!t]
  \centering
  \includegraphics[width=0.75\linewidth]{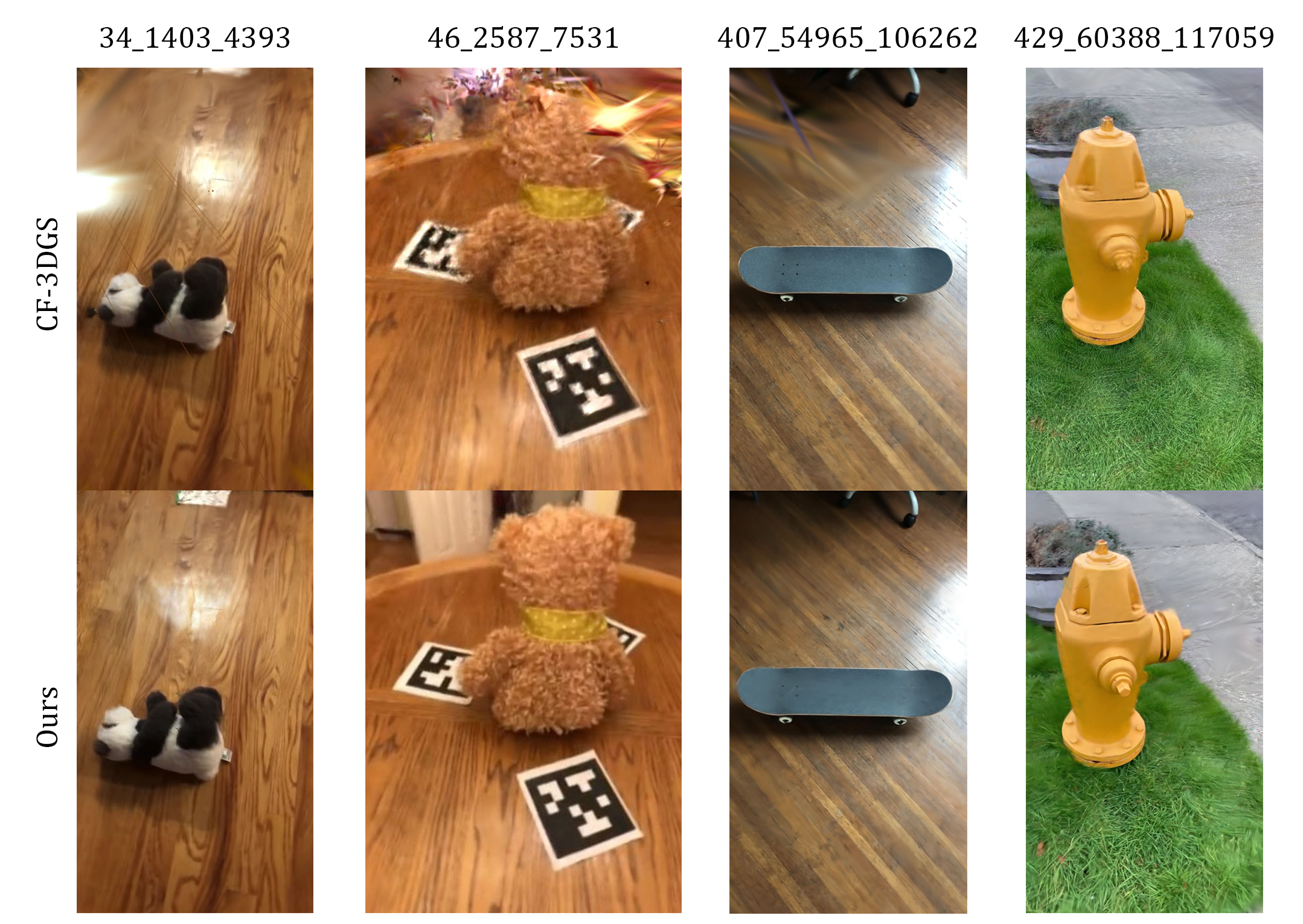}
   \caption{\textbf{Comparison of novel view synthesis on CO3D-V2.}}
   \label{fig:suppl-render-co3dv2}
\end{figure*}

As shown in Fig.~\ref{fig:suppl-render-T&T}, \ref{fig:suppl-render-synthetic} and ~\ref{fig:suppl-render-co3dv2}, which are evaluated by the same rules as mentioned in the main paper, our method outperforms other baselines by rendering more photo-realistic images, which benefits from the high quality rendering ability of 3DGS model and the accurate camera parameters estimated by our joint optimization. 
\red{Due to space limitations, we present only a subset of the results in the main text. The complete results for the Tanks and Temples and CO3D-V2 datasets are provided in Tab.~\ref{tab:suppl-nvs-TT-co3d}.}

\begin{table*}[!t]
    \centering
    \begingroup
    \footnotesize
    {
    \begin{tabular}{c c cccc cccc ccc} \hline
        & \multirow{2}{*}{Scenes} & \multicolumn{3}{c}{Ours}  & & \multicolumn{3}{c}{SFGS} & & \multicolumn{3}{c}{CF-3DGS} \\ \cline{3-5}\cline{7-9}\cline{11-13}
        &                         & PSNR$\uparrow$ & SSIM$\uparrow$ & LPIPS$\downarrow$ & & PSNR & SSIM & LPIPS & & PSNR & SSIM & LPIPS   \\ \hline
    \multirow{9}{*}{\rotatebox{90}{Tanks and Temples}}
        & Church                  & 25.56& 0.84& 0.16& & 22.46& 0.72&  0.22& & 21.96& 0.71& 0.27\\
        & Barn                    & 26.59& 0.86& 0.15& & 13.39& 0.53&  0.52& & 12.94& 0.53& 0.57\\
        & Museum                  & 24.50& 0.76& 0.22& & 18.68& 0.54&  0.43& & 16.08& 0.44& 0.52\\
        & Family                  & 27.03& 0.86& 0.15& & 15.92& 0.51&  0.51& & 15.99& 0.49& 0.53\\
        & Horse                   & 27.60& 0.90& 0.12& & 20.86& 0.75&  0.19& & 17.36& 0.64& 0.33\\
        & Ballroom                & 25.94& 0.86& 0.12& & 17.42& 0.49&  0.27& & 22.20& 0.72& 0.25\\
        & Francis                 & 30.09& 0.89& 0.18& & 27.22& 0.83&  0.20& & 17.72& 0.56& 0.48\\
        & Ignatius                & 22.12& 0.71& 0.22& & 12.16& 0.32&  0.72& & 14.96& 0.31& 0.59\\ \cline{2-13}
        & mean                    & \cellcolor{gray!40}26.18& \cellcolor{gray!40}0.84& \cellcolor{gray!40}0.17& & 18.51& 0.59&  0.38& & 17.40& 0.55& 0.44\\ \hline
     \multirow{9}{*}{\rotatebox{90}{CO3D-V2}}  
        & {34\_1403\_4393}          & 28.68 & 0.88 & 0.21   & & 32.52& 0.93& 0.14   & & 27.75 & 0.86 & 0.20         \\
        & {46\_2587\_7531}          & 31.83 & 0.92 & 0.12   & & 30.65& 0.91& 0.13   & & 25.44 & 0.80 & 0.21         \\
        & {106\_12648\_23157}       & 26.18 & 0.83 & 0.19   & & 23.43& 0.73& 0.28   & & 22.14 & 0.64 & 0.34         \\
        & {110\_13051\_23361}       & 33.44 & 0.94 & 0.11   & & 29.95& 0.87& 0.19   & & 29.69 & 0.89 & 0.29         \\
        & {245\_26182\_52130}       & 33.82 & 0.93 & 0.20   & & 28.59& 0.87& 0.27   & & 27.24 & 0.85 & 0.30         \\
        & {407\_54965\_106262}      & 28.73 & 0.86 & 0.35   & & 28.36& 0.85& 0.36   & & 27.80 & 0.84 & 0.35         \\
        & {415\_57112\_110099}      & 30.37 & 0.88 & 0.22   & & 27.23& 0.78& 0.30   & & 22.14 & 0.64 & 0.34        \\
        & {429\_60388\_117059}      & 25.70 & 0.70 & 0.35   & & 24.91& 0.70& 0.36   & & 24.44 & 0.68 & 0.36         \\ \cline{2-13}
        & mean                      & \cellcolor{gray!40}29.84 & \cellcolor{gray!40}0.87 & \cellcolor{gray!40}0.21   & & 28.73& 0.85& 0.22   & & 25.83 & 0.78 & 0.30        \\ \hline
    \end{tabular}
    }
    \caption{\textbf{NVS results on Tanks and Temples and CO3D-V2.} Each baseline method is trained with its public code under the original settings and evaluated with the same evaluation protocol.}
    \label{tab:suppl-nvs-TT-co3d}
    \endgroup
\end{table*}

\subsection{Camera Parameter Estimation}
\begin{figure*}[!t]
  \centering
  \includegraphics[width=\linewidth]{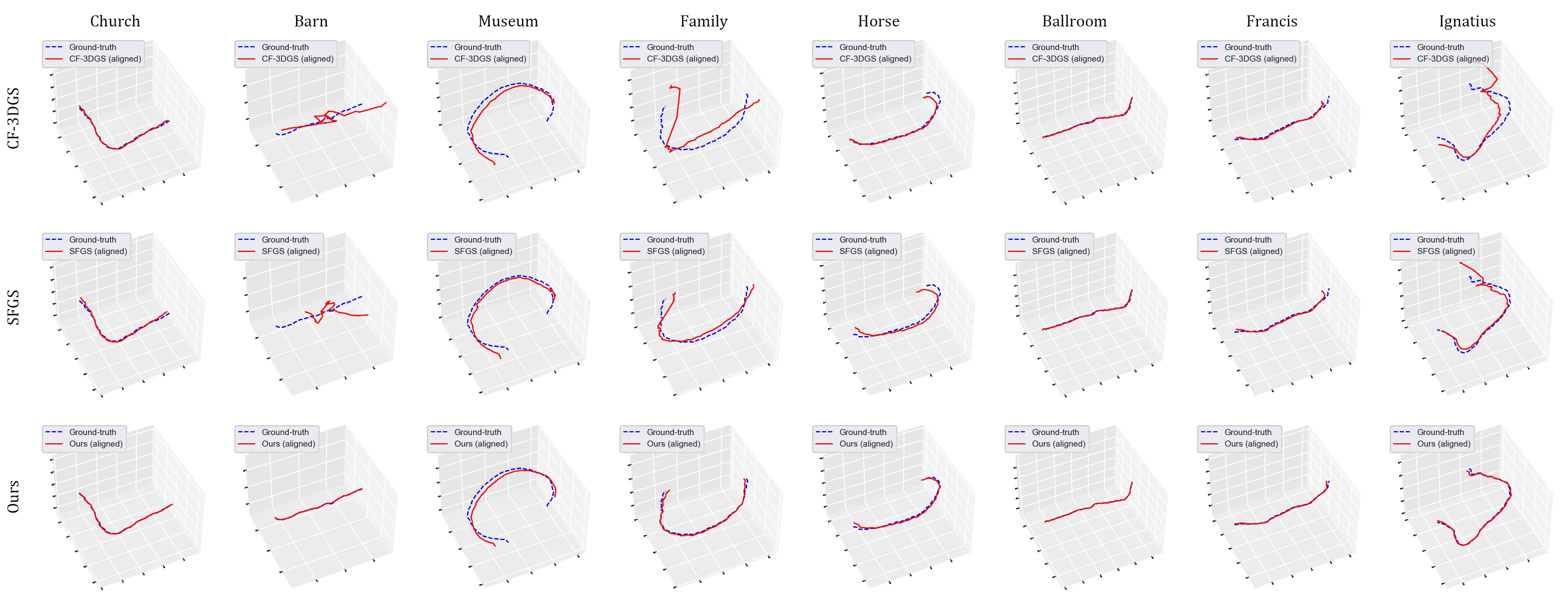}
   \caption{\textbf{Comparison of pose estimation on Tanks and Temples.}}
   \label{fig:suppl-pose-tt}
\end{figure*}

\begin{figure*}[t]
  \includegraphics[width=\linewidth]{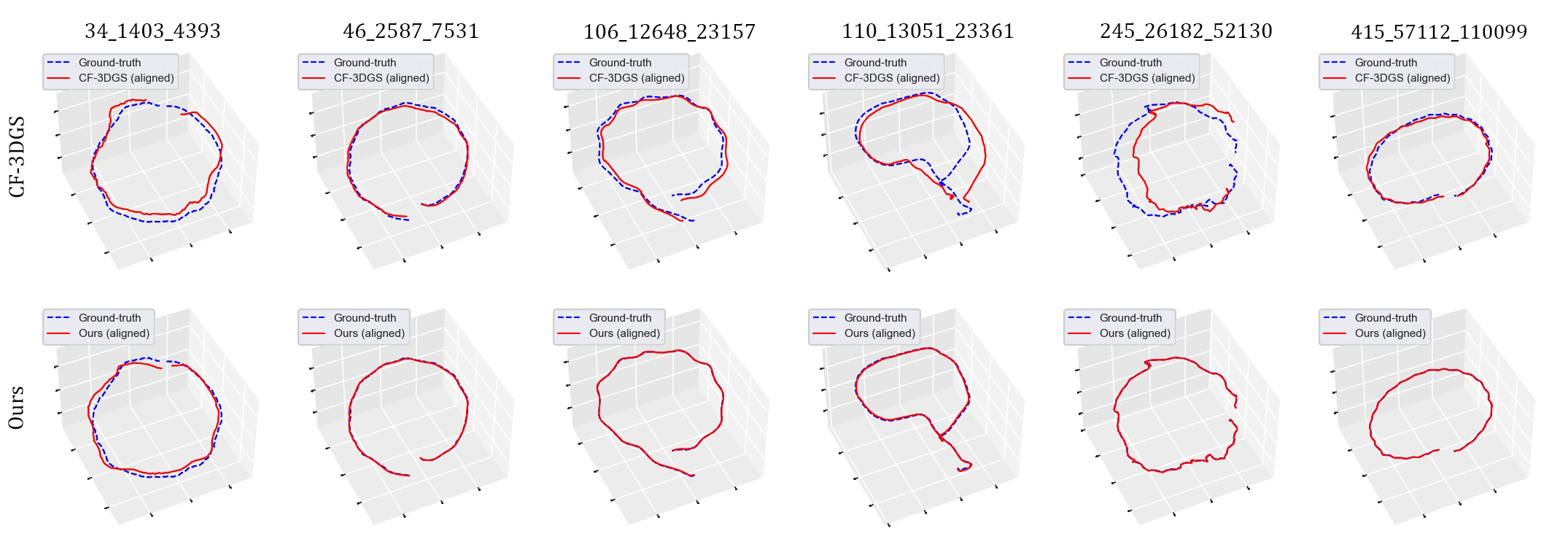}
   \caption{\textbf{Comparison of pose estimation on CO3D-V2.}}
   \label{fig:suppl-pose-co3dv2}
\end{figure*}

\red{
Fig.~\ref{fig:suppl-pose-tt}, \ref{fig:suppl-pose-synthetic} and \ref{fig:suppl-pose-co3dv2} illustrate the pose estimation results of our method and baseline approaches, compared against the ground-truth camera trajectories. As shown in Table~\ref{table:suppl-pose-T&T-co3dv2}, our method achieves 4–10× lower pose errors compared to the baselines, benefiting from our joint optimization framework that simultaneously refines camera parameters and 3DGS model under global track consistency constraints.
}

\begin{table*}[!t]
    \centering
    \begingroup
    \footnotesize
    {
    \begin{tabular}{c c cccc cccc ccc} \hline
        & \multirow{2}{*}{Scenes} & \multicolumn{3}{c}{Ours}  & & \multicolumn{3}{c}{SFGS} & & \multicolumn{3}{c}{CF-3DGS} \\ \cline{3-5}\cline{7-9}\cline{11-13}
        &                         & $\mathrm{RPE}_{t}\downarrow$ & $\mathrm{RPE}_{r}\downarrow$ & ATE$\downarrow$ & & $\mathrm{RPE}_{t}$ & $\mathrm{RPE}_{r}$ & ATE & & $\mathrm{RPE}_{t}$ & $\mathrm{RPE}_{r}$ & ATE  \\ \hline
    \multirow{9}{*}{\rotatebox{90}{Tanks and Temples}}
        & Church                  & 0.015& 0.039& 0.002& & 0.039& 0.172&  0.006& & 0.037& 0.190& 0.005 \\
        & Barn                    & 0.032& 0.131& 0.002& & 0.352& 0.865&  0.068& & 0.567& 0.620& 0.053 \\
        & Museum                  & 0.054& 0.356& 0.013& & 0.100& 0.452&  0.012& & 0.103& 0.676& 0.017 \\
        & Family                  & 0.010& 0.094& 0.003& & 0.662& 2.065&  0.036& & 0.590& 1.865& 0.031 \\
        & Horse                   & 0.018& 0.165& 0.006& & 0.054& 0.302&  0.015& & 0.098& 0.495& 0.012 \\
        & Ballroom                & 0.007& 0.018& 0.001& & 0.028& 0.062&  0.004& & 0.030& 0.100& 0.004 \\
        & Francis                 & 0.010& 0.157& 0.003& & 0.030& 0.320&  0.009& & 0.070& 0.509& 0.012 \\
        & Ignatius                & 0.015& 0.102& 0.003& & 0.191& 1.617&  0.016& & 0.211& 1.617& 0.025 \\ \cline{2-13}
        & mean                    & \cellcolor{gray!40}0.020& \cellcolor{gray!40}0.133& \cellcolor{gray!40}0.004& & 0.182& 0.732&  0.021& & 0.213& 0.759& 0.020 \\ \hline
     \multirow{9}{*}{\rotatebox{90}{CO3D-V2}}  
        & {34\_1403\_4393}          & 0.099 & 0.605 & 0.009     & & 0.041& 0.170& 0.009     & & 0.505 & 0.211 & 0.009         \\
        & {46\_2587\_7531}          & 0.013 & 0.080 & 0.001     & & 0.025& 0.275& 0.004     & & 0.095 & 0.447 & 0.009         \\
        & {106\_12648\_23157}       & 0.009 & 0.076 & 0.001     & & 0.045& 0.282& 0.014     & & 0.094 & 0.360 & 0.008         \\
        & {110\_13051\_23361}       & 0.012 & 0.052 & 0.001     & & 0.093& 0.331& 0.020     & & 0.140 & 0.401 & 0.021        \\
        & {245\_26182\_52130}       & 0.005 & 0.029 & 0.001     & & 0.064& 0.438& 0.017     & & 0.239 & 0.472 & 0.017         \\
        & {407\_54965\_106262}      & 0.062 & 0.461 & 0.011     & & 0.122& 0.566& 0.018     & & 0.310 & 0.243 & 0.008         \\
        & {415\_57112\_110099}      & 0.004 & 0.024 & 0.001     & & 0.049& 0.351& 0.024     & & 0.110 & 0.424 & 0.014         \\
        & {429\_60388\_117059}      & 0.052 & 0.454 & 0.009     & & 0.081& 0.412& 0.027     & & 0.134 & 0.542 & 0.018         \\ \cline{2-13}
        & mean                      & \cellcolor{gray!40}0.032 & \cellcolor{gray!40}0.222 & \cellcolor{gray!40}0.004     & & 0.053& 0.308& 0.017     & & 0.203 & 0.388 & 0.013         \\ \hline
    \end{tabular}
    }
    \caption{\textbf{Quantitative comparison of pose accuracy on Tanks and Temples and CO3D-V2.} The unit of $\mathrm{RPE}_{r}$ is in degrees, ATE is in the ground truth scale and $\mathrm{RPE}_{t}$ is scaled by 100. 
    }
    \label{table:suppl-pose-T&T-co3dv2}
    \endgroup
\end{table*}

\subsection{Rendering Trajectory}
\begin{figure*}[!t]
  \centering
  \includegraphics[width=\linewidth]{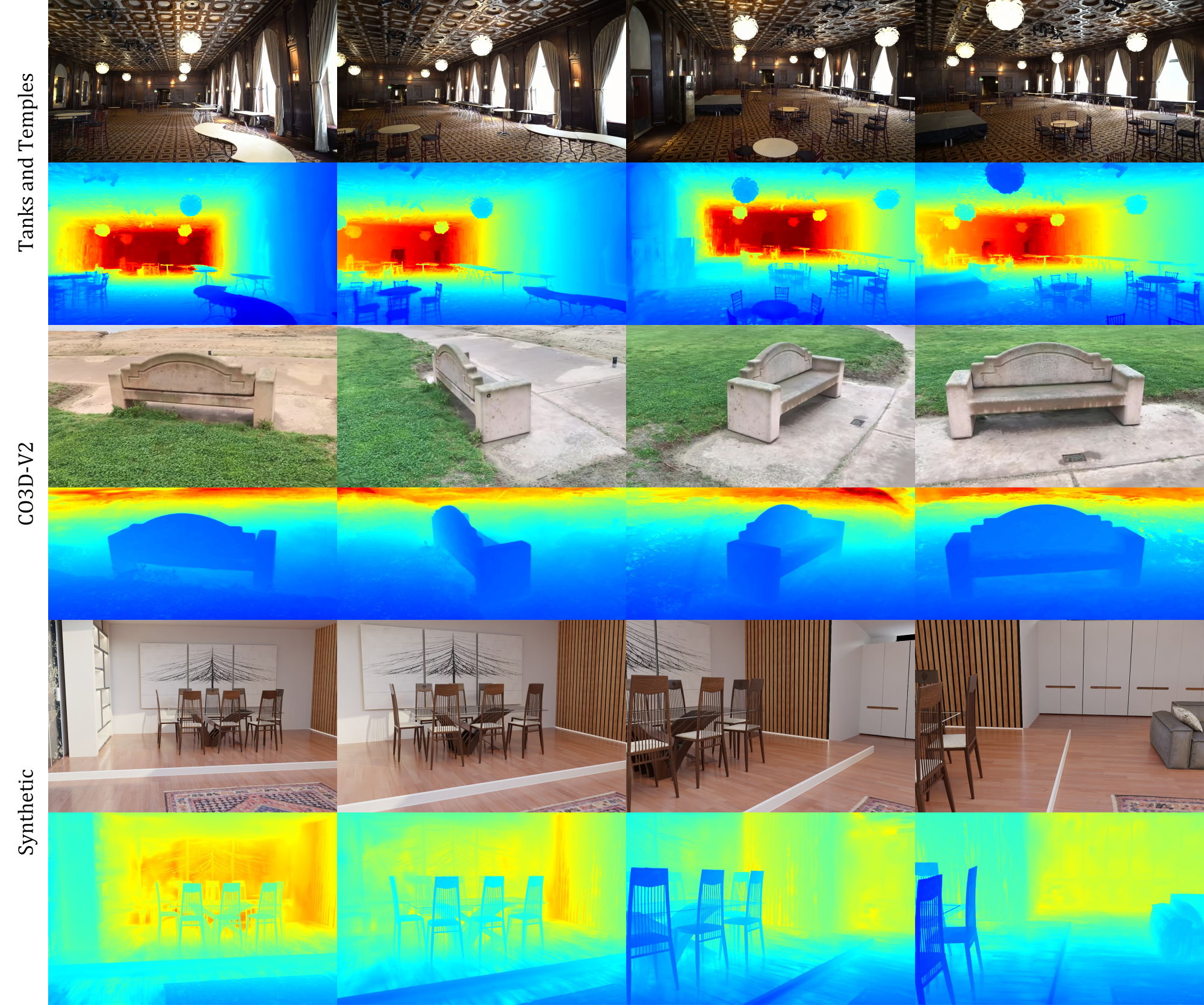}
   \caption{\textbf{Novel view synthesis on a new camera trajectory.} We demonstrate the rendering images and the associated depth maps on scenes from three datasets, where the view points are uniformly sampled on a \textbf{new} camera trajectory.}
   \label{fig:suppl-render-nvs}
\end{figure*}

To better illustrate the results of novel view synthesis, we have also created several videos showcasing continuous camera motion using these datasets. Fig.~\ref{fig:suppl-render-nvs} shows novel view synthesis results on scenes from the datasets, in which novel views are sampled from new camera trajectories.

\subsection{Ablation Study}
To exhibit the effectiveness of different losses in our joint optimization, we ablate each loss of the algorithm on synthetic dataset, since it has ground-truth camera parameters. Tab.~\ref{table:suppl-abla-details} reports the synthesis quality and camera parameter errors accross different variants on our synthetic dataset. 

\begin{table*}[!ht]
    \centering
    \resizebox{\linewidth}{!}{
    \begin{tabular}{c|ccccc|ccccc|ccccc|ccccc} \hline
    \multirow{2}{*}{Scene}  & \multicolumn{5}{c}{classroom}         & \multicolumn{5}{c}{{lego\_c2}}        & \multicolumn{5}{c}{livingroom}        & \multicolumn{5}{c}{bedroom} \\ \cline{2-21}
                            & PSNR  & SSIM  & LPIPS & ATE   & FoV           & PSNR  & SSIM  & LPIPS & ATE   & FoV           & PSNR  & SSIM  & LPIPS & ATE   & FoV               & PSNR  & SSIM  & LPIPS & ATE   & FoV \\ \hline
    COLMAP+3DGS             &   35.81&   0.94&   0.15&   0.00023&   0.993   &   28.77&   0.88&   0.15&   0.00019&   0.021   &   32.74&   0.87&   0.27&   0.00014&   0.0291      &   31.73&   0.94&   0.13&   0.00023&  0.042\\
    w.o. 2D Track Loss      &   24.08&   0.78&   0.40&   0.00973&   1.668   &   17.60&   0.35&   0.42&   0.01969&   3.013   &   20.82&   0.62&   0.45&   0.01735&   3.3760      &   10.22&   0.50&   0.60&   0.03404&  2.413\\
    w.o. 3D Track Loss      &   35.99&   0.94&   0.14&   0.00012&   0.083   &   29.33&   0.90&   0.13&   0.00012&   0.031   &   33.24&   0.88&   0.26&   0.00021&   0.1150      &   30.97&   0.93&   0.14&   0.01056&  0.021\\
    Ours                    &   36.26&   0.95&   0.13&   0.00008&   0.012   &   29.36&   0.90&   0.12&   0.00011&   0.031   &   33.52&   0.88&   0.24&   0.00009&   0.0121      &   31.17&   0.93&   0.13&   0.00044&  0.003\\ \hline
    \end{tabular}
    }
    \caption{\textbf{Ablatation study on different losses on Synthetic Dataset.} }
    \label{table:suppl-abla-details}
\end{table*}

\bibliography{aaai2026}